\documentclass[letterpaper]{article} 
\usepackage{aaai2027}  
\usepackage[hyphens]{url}  
\usepackage{graphicx} 
\usepackage{natbib}  
\usepackage{caption} 
\usepackage{amsmath,amssymb}
\usepackage{algorithm}
\usepackage{algorithmic}
\usepackage{newfloat}
\usepackage{listings}
\nocopyright

\DeclareCaptionStyle{ruled}{labelfont=normalfont,labelsep=colon,strut=off} 
\floatstyle{ruled}
\newfloat{listing}{tb}{lst}{}
\floatname{listing}{Listing}
\usepackage{booktabs}
\usepackage{colortbl}
\usepackage{xspace}
\usepackage{multicol}
\definecolor{groupgray}{gray}{0.92}
\definecolor{motivationgray}{gray}{0.96}
\definecolor{motivationrule}{gray}{0.62}
\definecolor{candidateRed}{RGB}{0,0,0}
\definecolor{wjtDeepBlue}{RGB}{20,55,120}

\newcommand{\sys}{ExFold\xspace}

\title{\sys: Unified Expert Folding for Training-Free MoE Prefill-Decode Acceleration}

\author{
    Juntong Wu \textsuperscript{\rm 1, 2}\equalcontrib ,
    Yifei Liu \textsuperscript{\rm 1, 3}\equalcontrib ,
    Junyi Chen \textsuperscript{\rm 1, 3}, 
    Siqi Fan \textsuperscript{\rm 1, 4},
    Chaoran Feng \textsuperscript{\rm 2} \\ \vspace{1mm}
    Minghao Li \textsuperscript{\rm 1},
    Liujie Zhang \textsuperscript{\rm 1},
    Weihang Cheng \textsuperscript{\rm 1}\corresponding,
    Li Yuan \textsuperscript{\rm 2}\corresponding 
}
\affiliations{
    \textsuperscript{\rm 1} Xiaohongshu Inc. \\ \vspace{0.5mm}
    \textsuperscript{\rm 2} Shenzhen Graduate School, Peking University \\ \vspace{0.5mm}
    \textsuperscript{\rm 3} Shanghai Jiao Tong University \\ \vspace{0.5mm}
    \textsuperscript{\rm 4} University of Electronic Science and Technology of China \\ \vspace{0.5mm}
    \small{\textbf{Correspondence:} {chenjinzhi@xiaohongshu.com, yuanli-ece@pku.edu.cn}}
}

\begin{document}

    \maketitle

    \begin{abstract}

    Mixture-of-Experts (MoE) models scale capacity for strong quality while keeping per-token compute bounded through sparse expert activation.
    Yet low-latency MoE serving is increasingly challenging, because it spans two inference phases with fundamentally different bottlenecks: prefill is dominated by token-wise expert computation, whereas decode is constrained by memory traffic from the batch-wise activated expert set.
    However, existing training-free acceleration methods optimize only a single resource proxy---either the experts each token executes or the experts a batch activates---and, either discard the excluded experts' contribution or leave it only implicitly approximated.
    In this paper, we propose \sys, a unified training-free expert-folding framework for jointly accelerating MoE prefill and decode.
    \sys casts both prefill and decode as one budgeted output-approximation problem: execute only a phase-specific constrained expert set while projecting the contribution of budget-excluded experts onto retained experts using calibrated scalar projectors.
    Motivated by the observation that many expert outputs are directionally aligned but differ in magnitude, \sys calibrates a pairwise scalar-projector matrix on unlabeled data and uses it at inference time to fold excluded expert contributions into retained experts.
    Under this view, prefill acceleration becomes token-level Top-$K$ folding, and decode acceleration becomes batch-level expert-pool folding.
    The two phases differ only in how retained experts are selected, while excluded contributions are recovered by one shared folding mechanism.
    We implement \sys as a plug-and-play plugin in vLLM, with a lightweight expert-folding CUDA kernel, delivering up to 1.41$\times$ TTFT and 2.45$\times$ TPOT speedups while retaining about 99\% of the original average quality. Code of ExFold can be seen in \url{https://github.com/Time-Rune/ExFold-MoE}.
\end{abstract}

    \section{Introduction}

\begin{figure}[!ht]
    \centering
    \hspace{-0.05\columnwidth}\includegraphics[width=0.96\columnwidth]{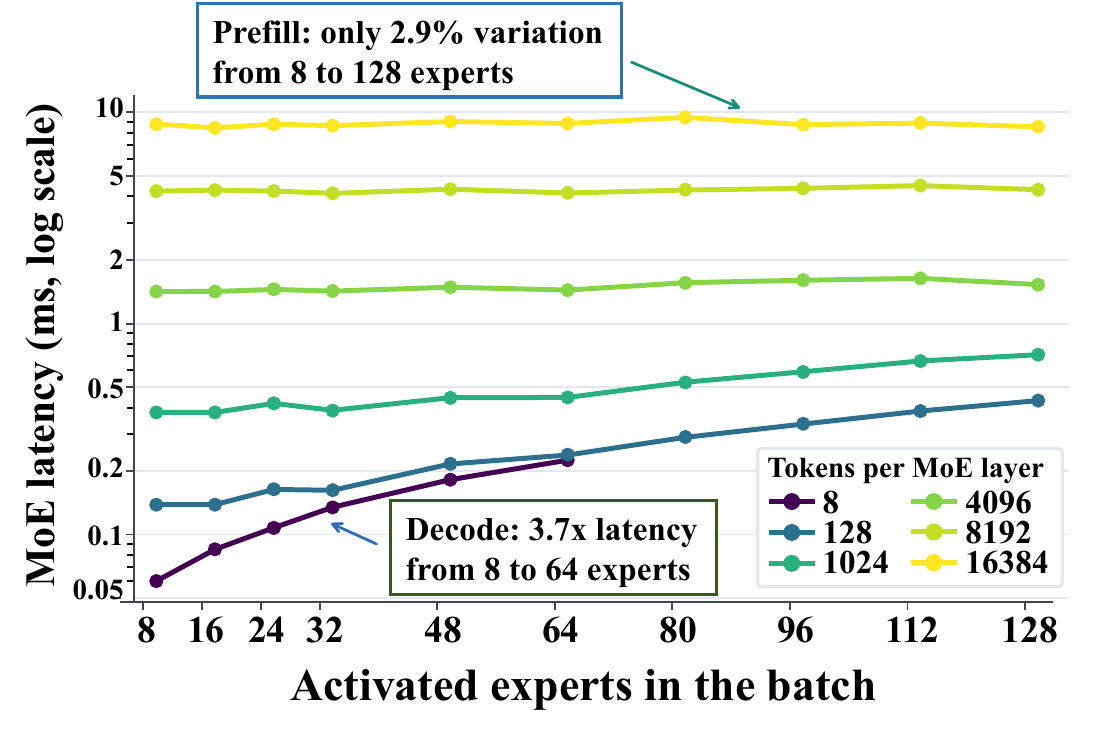}
    \caption{MoE prefill and decode have different bottlenecks.}
    \label{fig:placeholder}
\end{figure}

\begin{figure}[!ht]
    \centering
    \hspace{-0.03\columnwidth}
    \includegraphics[width=1.0\columnwidth]{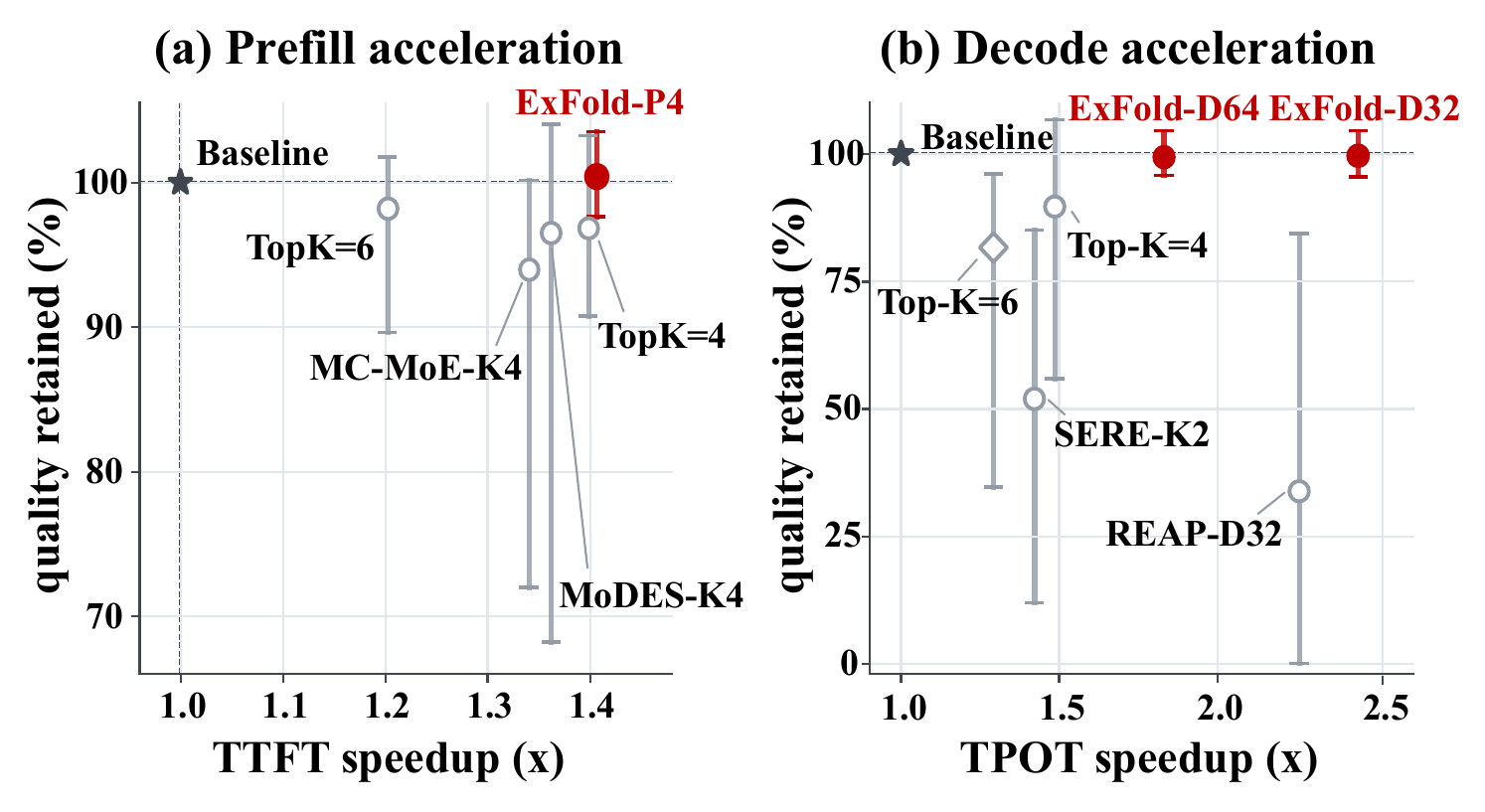}
    \caption{Quality--speed trade-offs for \sys prefill and decode acceleration.}
    \vspace{-2mm}
    \label{fig:method_result_overview}
    \label{fig:prefill_decode_bottleneck}
\end{figure}

Mixture-of-Experts (MoE) scales model parameters by orders of magnitude while keeping the compute budget bounded by activating only a sparse subset of parameters per token~\cite{fedus2021switch,jiang2024mixtral}.
This property has made MoE the mainstream design of choice for many recent large models seeking higher capability~\cite{jiang2024mixtral,qwen3a3b,glm45air} .
At the same time, it places stringent demands on low-latency serving~\cite{chen2026tokenflow, chen2025pre3, du2026c} in production: both the prefill latency (time-to-first-token, TTFT) and the decode latency (time-per-output-token, TPOT) matter, and the two exhibit distinct computational characteristics.
As shown in Fig.~\ref{fig:placeholder}, prefill is dominated by token-level expert computation, whereas decode is dominated by batch-level expert memory traffic.


Existing training-free MoE acceleration methods fall into two families, according to which of these two budgets they optimize.
\emph{Token-wise expert sparsification} reduces the number of experts each token executes, including Dynamic-MoE~\cite{huang2024harder}, MoDES~\cite{huang2025modes}, NAEE~\cite{notall2024experts}, and MC-MoE~\cite{mcmoe2024}.
\emph{Expert-set consolidation} instead shrinks the active expert set, through static pruning and merging such as REAP~\cite{lasby2025reap}, HC-SMoE~\cite{chen2025hcsmoe}, and Sub-MoE~\cite{submoe2025}, and dynamic batch-level restriction such as Lynx~\cite{gupta2024lynx} and SERE~\cite{wu2026sere}.

Yet both families frame acceleration as a resource-reduction problem: \textbf{they answer only which experts to execute, and leave the harder question unaddressed---what becomes of the experts they cut}.
Their treatment of the excluded contribution is, at best, incidental.
Pruning and skipping discard it outright~\cite{notall2024experts}.
Static merging bakes it into a single, permanently compressed model that can no longer adapt to the per-token or per-batch budget it actually faces at inference~\cite{mcmoe2024}.
Similarity-based re-routing swaps an excluded expert for a nearby one, but never calibrates how far the substitute's output strays from the contribution it replaces~\cite{wu2026sere}.
Lost, frozen, or only implicitly approximated, the excluded expert mass is never explicitly reconstructed---so the approximation error compounds as the budget is tightest and these methods are pushed hardest.

We introduce \emph{Expert Folding}, which projects every budget-excluded expert contribution onto an executed expert instead of discarding it, and build \sys around it.
This design follows two empirical observations (Figure~\ref{fig:expert_output_alignment}): many source experts have at least one target with directionally aligned outputs, while output magnitudes differ substantially across experts.
The former makes expert substitution possible; the latter explains why direct re-routing is insufficient and motivates a directed scalar projector that corrects the source--target scale mismatch.
Concretely, a single frozen forward pass over unlabeled text calibrates two per-layer matrices---a \emph{scalar-projector matrix} and a \emph{projection-loss matrix}: at inference, the loss matrix routes each excluded expert to its minimum-loss target, and the scalar matrix folds each excluded contribution into that target's router weight.

Since folding is defined at the level of individual excluded experts, it applies unchanged in both phases.
Prefill selects $K_{\mathrm{pre}}$ dominant experts per token, and decode selects $D$ dominant experts per batch---both reuse the same scalar matrix, loss matrix, and folding operator to recover whatever falls outside the retained set.
This is the central design of \sys: the two phases differ only in how retained experts are selected, while excluded contributions are recovered by one shared mechanism.

We implement \sys as a plug-and-play plugin in vLLM, with a lightweight expert-folding CUDA kernel, delivering up to 1.41$\times$ TTFT and 2.45$\times$ TPOT speedups while retaining about 99\% of the original average quality.
Our contributions are as follows:
\begin{itemize}
    \item \textbf{Phase-aware formulation.} We characterize the distinct expert budgets of prefill and decode and unify both as a single output-approximation problem under phase-specific selection constraints.
    \item \textbf{Expert Folding.} We propose a training-free folding mechanism that projects budget-excluded expert contributions onto retained experts via a directional projector and a reconstruction-loss matrix, whose scalar form is absorbed directly into the router weight.
    \item \textbf{Evaluation.} We validate near-lossless prefill, decode, and joint acceleration across multiple MoE architectures on real vLLM serving workloads.
\end{itemize}

    \section{Background and Motivation}

\subsection{MoE Inference}

\begin{figure*}[!t]
    \centering
    \includegraphics[width=0.98\textwidth]{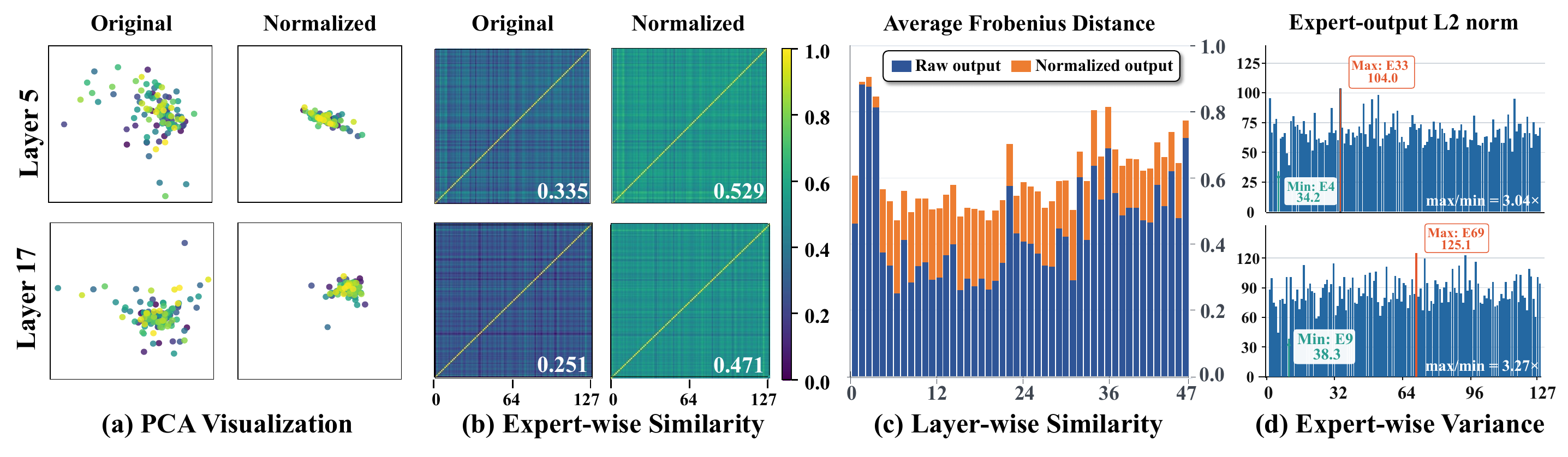}
    \caption{(a-c) MoE expert outputs can be aligned across experts and layers, and (d) their L2 norms vary substantially .}
    \label{fig:expert_output_alignment}
\end{figure*}

\paragraph{Sparse MoE layers.}
A sparse MoE layer replaces the dense FFN of a Transformer block with $N$ gated-FFN experts and a router~\cite{fedus2021switch,jiang2024mixtral}.
For a hidden state $x$ of dimension $d$, expert $e$ computes
\begin{equation}
   E_e(x) =
   W^{(e)}_{\mathrm{down}}
   \left(
       \phi\!\left(W^{(e)}_{\mathrm{gate}} x\right)
       \odot
       W^{(e)}_{\mathrm{up}} x
   \right),
   \label{eq:gated_expert}
\end{equation}
where $\phi(\cdot)$ is typically SiLU and $\odot$ denotes element-wise multiplication.
The router produces logits $r(x)=W_r x$ over $N$ routed experts, selects the Top-$K$ set $S_K(x)$, and normalizes the gate weights within it:
\begin{equation}
   \alpha_e(x) =
   \frac{\exp(r_e(x))}
   {\sum_{j \in S_K(x)} \exp(r_j(x))},
   \quad e \in S_K(x).
   \label{eq:topk_gate}
\end{equation}
The MoE output is then
\begin{equation}
   \mathrm{MoE}(x)=
   \sum_{e \in S_K(x)} \alpha_e(x) E_e(x),
   \label{eq:moe_output}
\end{equation}
possibly with additional shared experts.
This computation makes the FFN path sparse per token, but it also introduces dynamic routing, expert dispatch, and expert-output aggregation.

\paragraph{Phase-specific budgets.}
Although the same MoE layer (Eq.~\ref{eq:moe_output}) runs in both phases, its bottleneck shifts with the shape of the token batch (Figure~\ref{fig:placeholder}).
Prefill processes a prompt of $T$ tokens at once, and each token independently activates its own support $S_K(x)$, so the layer performs $\mathcal{O}(TK)$ expert-FFN evaluations.
When $T$ is large, every loaded expert is reused across many tokens, so weight-loading is amortized and the phase is compute-bound on token-wise expert FLOPs; reducing $K$ directly cuts this cost.
Decode advances each of $B$ batched requests by a single token per step, issuing only $B$ tokens, yet their supports $\bigcup_x S_K(x)$ typically cover most of the $N$ experts.
The layer must therefore load this batch-wise expert union to serve very few tokens each, yielding low arithmetic intensity and a memory-bound phase whose cost tracks the number of distinct activated experts rather than FLOPs.
Efficient serving is thus governed by two different budgets: experts per token in prefill, and experts per batch in decode.

\subsection{MoE Acceleration Methods}

Depending on which budget they optimize, existing training-free MoE acceleration methods fall into two families: token-wise expert sparsification and expert-set consolidation.

\paragraph{Token-wise expert sparsification.}
This family reduces the experts each token executes.
Dynamic-MoE keeps per token the smallest expert set with cumulative router probability above $p$, so easier tokens use fewer experts~\cite{huang2024harder}.
MoDES skips experts per token by a calibration-estimated importance~\cite{huang2025modes}, NAEE prunes low-importance executions from router or activation statistics~\cite{notall2024experts}, and MC-MoE merges co-activated experts identified from calibration~\cite{mcmoe2024}.
These methods relieve the prefill computation budget, but leave the decode-time expert union uncontrolled.

\paragraph{Expert-set consolidation.}
This family shrinks the active expert set.
Static methods score every expert from calibration signals and permanently reduce the global pool: REAP prunes the least important experts by router weight and activation norm~\cite{lasby2025reap}, while HC-SMoE~\cite{chen2025hcsmoe}, REAM~\cite{jha2026ream}, and Sub-MoE~\cite{submoe2025} merge similar experts.
Dynamic methods restrict the expert union within a decoding batch: Lynx re-routes secondary requests to active experts~\cite{gupta2024lynx}, and SERE calibrates pairwise expert similarity offline and substitutes each excluded expert with a similar retained one at decode~\cite{wu2026sere}.
These methods relieve decode-side traffic, but their static transformations do not match the per-token prefill budget.

\paragraph{Gap.}
Both families decide \emph{which} experts to execute, but neither preserves the contribution of the experts they exclude: pruning discards it, static merging bakes it into a permanently compressed model, and similarity re-routing changes an expert's destination without calibrating how far the substitute strays.
As the budget tightens, this error grows within each phase and compounds across prefill and decode, which are approximated against two different targets rather than one shared objective.

\begin{center}
    \setlength{\fboxsep}{6pt}
    \fcolorbox{motivationrule}{blue!5}{\begin{minipage}{0.93\columnwidth}
    \renewcommand{\baselinestretch}{1.10}\selectfont
    \noindent\textbf{Key Question}\\[2pt]
    \emph{How can we design a unified framework to accelerate the prefill and decode of MoE models with minimal loss?}
    \end{minipage}}
\end{center}

    \section{Design Insight}

The gap makes recovery look hard from both sides---prefill and decode pull toward conflicting budgets, and faithfully restoring each excluded expert seems to require either recomputation or blind substitution---yet two observations dissolve both difficulties.

\paragraph{The excluded contributions share one recovery target.}
Prefill and decode are constrained by different budgets---experts per token versus experts per batch---yet whichever experts a budget excludes, the residual it leaves behind has the identical form $\alpha_e(x)E_e(x)$ in the MoE output (Eq.~\ref{eq:moe_output}).
The two phases therefore pose a single problem---reconstruct the excluded contributions from the retained experts---and differ only in the constraint that selects the retained set, not in what must be recovered.

\paragraph{Expert redundancy is magnitude-separable.}
Figure~\ref{fig:expert_output_alignment} shows that the redundancy across experts is concentrated in one cheaply correctable degree of freedom---output scale.
In a two-component PCA projection (a), raw expert outputs form a diffuse cloud, but normalizing each to unit scale collapses them onto a thin shared axis: once scale is removed, experts point in nearly the same direction.
Pairwise similarity confirms this---normalization lifts the average expert-to-expert similarity from $0.335$ to $0.529$ at layer~5 and from $0.251$ to $0.471$ at layer~17 (b)---and the same gap persists across all $48$ layers (c), so the alignment is structural rather than a few-layer artifact.
Yet the discarded scale is large: raw output norms span over $3\times$ within a layer ($\max/\min=3.04\times$ and $3.27\times$ for the two shown, d).
The mismatch between an excluded and a retained expert is therefore almost purely radial---a compatible direction is already available, only its magnitude is off.
A single per-pair scalar that rescales the retained expert---not a reconstructed vector or a blind substitution---thus bounds how far the substitute strays and makes faithful recovery cheap in principle.
\textcolor{candidateRed}{Full-layer visualization is provided in Appendix D.}

Together these give the motivating question an affirmative answer: one recovery target, reached by a cheap per-pair scalar correction and shared across both phases, with each phase changing only how the retained set is selected. \sys turns this into a concrete mechanism in the next section.

    \begin{figure*}[!ht]
    \centering
    \includegraphics[width=.99\textwidth]{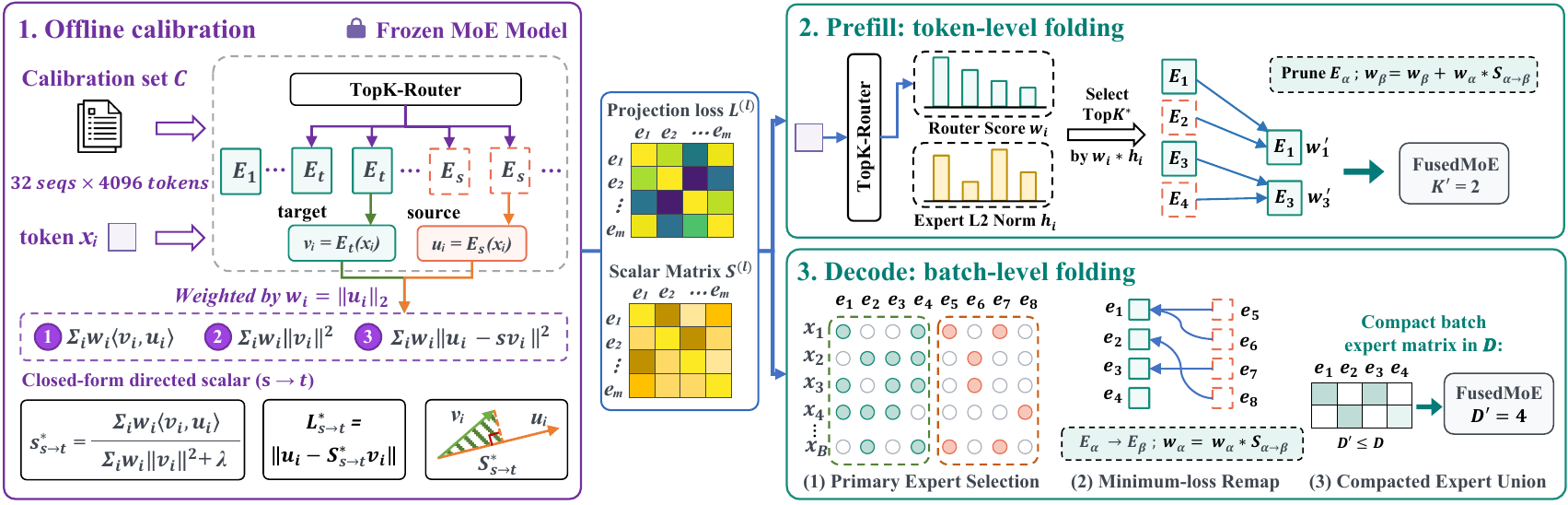}
    \caption{Overview of \textbf{\sys}: training-free projector calibration and unified expert folding.}
    \vspace{-4mm}
    \label{fig:uniep_method}
\end{figure*}

\section{Method}

\subsection{Unified Expert Folding}

\sys treats MoE acceleration as constrained output approximation rather than phase-specific expert dropping. Let $x_i$ denote the $i$-th input token and $S_K(x_i)$ its original Top-$K$ support. We extend the router weight by $\alpha_e(x_i)=0$ for $e\notin S_K(x_i)$. Under phase $\phi\in\{\mathrm{pre},\mathrm{dec}\}$, $B_\phi(x_i)$ denotes the retained target set and $O_\phi(x_i)=S_K(x_i)\setminus B_\phi(x_i)$ the omitted source experts. We use $E_s$ for an omitted source expert and $E_t$ for its retained target, with $t=\pi_\phi(s,x_i)\in B_\phi(x_i)$.

Let $s_{s\to t}^{\star}$ be a directed scalar projector such that $E_s(x_i)\approx s_{s\to t}^{\star}E_t(x_i)$, and write $t_s=\pi_\phi(s,x_i)$ for the target assigned to source $s$. The approximated MoE output separates the retained contributions from the projected contributions of omitted experts:
\begin{equation}
    \begin{aligned}
    \widehat{y}_\phi(x_i)
    &=
    \underbrace{
    \sum_{t\in S_K(x_i)\cap B_\phi(x_i)}
    \alpha_t(x_i)E_t(x_i)
    }_{\text{retained}}\\
    &\quad+
    \underbrace{
    \sum_{s\in O_\phi(x_i)}
    \alpha_s(x_i)s_{s\to t_s}^{\star}E_{t_s}(x_i)
    }_{\text{folded}} .
    \end{aligned}
    \label{eq:folded_moe}
\end{equation}
Thus, omitted experts are not executed, while their calibrated contributions remain in the MoE output. Routes folded into the same target are coalesced before expert computation, so each retained target is evaluated only once. Unlike static expert merging, folding changes neither expert parameters nor the original router~\cite{mcmoe2024,chen2025hcsmoe}.

For the tokens $\mathcal{X}_\phi$ processed together in one layer, both phases share the objective
\begin{equation}
    \min_{B_\phi,\pi_\phi,\mathbf{S}}
    \sum_{x_i\in\mathcal{X}_\phi}
    \|\mathrm{MoE}(x_i)-\widehat{y}_\phi(x_i)\|_2^2
    \quad\mathrm{s.t.}\quad
    c_\phi(B_\phi)\leq\tau_\phi,
    \label{eq:unified_budget}
\end{equation}
where $\mathbf{S}=[s_{s\to t}^{\star}]$ is the scalar-projector table. The prefill cost counts retained experts per token, whereas the decode cost counts distinct experts retained for a batch. We next calibrate $\mathbf{S}$ once and then solve the two budget constraints with the same folding rule.

\subsection{Training-Free Projector Calibration}

Figure~\ref{fig:uniep_method} summarizes the offline calibration. For each ordered, co-routed source-target pair $(s,t)$, we collect their outputs on $m$ calibration tokens:
\begin{equation}
    \mathbf{u}_i=E_s(x_i),\qquad
    \mathbf{v}_i=E_t(x_i),\qquad i=1,\ldots,m.
    \label{eq:calibration_outputs}
\end{equation}
Here $\mathbf{u}_i,\mathbf{v}_i\in\mathbb{R}^{d}$, and each token is weighted by $w_i=\|\mathbf{u}_i\|_2$ in the main method. Given a projector family $\mathcal{P}$, calibration solves the weighted output-reconstruction problem
\begin{equation}
    \mathbf{P}_{s\to t}^{\star}
    =
    \arg\min_{\mathbf{P}\in\mathcal{P}}
    \sum_{i=1}^{m}w_i
    \|\mathbf{u}_i-\mathbf{v}_i\mathbf{P}\|_2^2
    +\lambda\Omega(\mathbf{P}),
    \label{eq:projector_objective}
\end{equation}
where $\Omega$ regularizes the parameters of the selected projector family. For the scalar parameterization $\mathbf{P}=s\mathbf{I}$, the solution is
\begin{equation}
    s_{s\to t}^{\star}
    =
    \frac{\sum_{i=1}^{m}w_i
    \langle\mathbf{v}_i,\mathbf{u}_i\rangle}
    {\sum_{i=1}^{m}w_i
    \|\mathbf{v}_i\|_2^2+\lambda}.
    \label{eq:scalar_projector}
\end{equation}

Besides the scalar form, we evaluate diagonal and scalar-plus-low-rank projectors in Table~\ref{tab:prefill_quick_results}.
The deployed method uses the scalar solution because it can be folded into router weights without an online vector transform.
\textcolor{candidateRed}{We specify these alternatives, parameterization, and closed-form calibration in Appendix B.}

Finally, scalar transfer error is stored as
\begin{equation}
    \ell_{s\to t}
    =\frac{\sum_{i=1}^{m}w_i
      \|\mathbf{u}_i-s_{s\to t}^{\star}\mathbf{v}_i\|_2^2}
    {\sum_{i=1}^{m}w_i\|\mathbf{u}_i\|_2^2}.
    \label{eq:projection_loss}
\end{equation}
Calibration therefore returns one directed scalar table $\mathbf{S}^{(l)}=[s_{s\to t}^{\star}]$ and loss table $\mathbf{L}^{(l)}=[\ell_{s\to t}]$ per MoE layer, without labels, gradients, or model updates. Detailed settings, matrix, and collections are provided in Appendices A and D.

\begin{table*}[!t]
    \centering
    \small
    \setlength{\tabcolsep}{3pt}
    \resizebox{.96\textwidth}{!}{
    \begin{tabular}{l|cc|cccccccc|c}
        \toprule
        \rowcolor{gray!30}
        \textbf{Method}                                 & \textbf{Prefill}                        & \textbf{Decode}                           & \textbf{MATH$_{500}$}                         & \textbf{AIME24}                          & \textbf{IFEval}                          & \textbf{IFBench}                         & \textbf{GPQA}                            & \textbf{LCB (P/A.@8)}                                     & \textbf{Eval+}                           & \textbf{MMLU$_{pro}$}   & \textbf{Avg.}                 \\
        \midrule
        \rowcolor{groupgray}\multicolumn{12}{l}{\textbf{Baseline}} \\
        {Original TopK=8}                         & 100\%                          & 100\%                            & 97.40                           & 64.48                           & 83.73                           & 29.31                           & 63.13                           & 68.26 / 56.89                           & 77.44                           & 68.57                           & 69.04 \\
        \cmidrule(lr){1-12}
        \rowcolor{groupgray}\multicolumn{12}{l}{\textbf{Prefill-Only Acceleration}} \\
        Prefill TopK=6           & 75\%                           & 100\%                            & 95.80                           & 64.79                           & 83.18                           & 26.27                           & 64.14                           & 69.46 / 57.34                           & 75.61                           & 66.33                           & 68.20 \\
        Prefill TopK=4           & 50\%                           & 100\%                            & 96.00                           & 66.56                           & 76.00                           & 29.02                           & 60.61                           & 66.47 / 57.63                           & 73.17                           & 65.28                           & 66.64 \\
        {MC-MoE-P4}    & {50\%} & {100\%} & {96.00} & {64.58} & {79.85} & {26.48} & {45.45} & {68.26 / 57.11} & {77.44} & {65.44} & {65.44} \\
        {MoDES-P4}     & {50\%} & {100\%} & {95.40} & \textbf{67.08} & {82.26} & \textbf{30.26} & {43.06} & {69.46 / 57.71} & \textbf{78.66} & {66.53} & {66.59} \\
        \rowcolor{blue!5}
        \textbf{\sys-P4}                       & 50\%                           & 100\%                            & \textbf{97.00}                           & 66.46                           & \textbf{84.00}                           & 29.53                           & \textbf{62.63}                           & \textbf{70.66} / \textbf{57.86}                           & 76.83                           & \textbf{66.94}                           & \textbf{69.26} \\
        \cmidrule(lr){1-12}
        \rowcolor{groupgray}\multicolumn{12}{l}{\textbf{Decode-Only Acceleration}} \\
        REAP-D64                               & 100\%                          & 50.0\%                           & 94.80                           & 65.73                           & 71.16                           & \textbf{31.29}                           & 35.35                           & 67.07 / 52.99                           & 75.00                           & 51.61                           & 61.50 \\
        REAP-D32                               & 100\%                          & 25.0\%                           & 70.00                           & 23.02                           & 33.83                           & 24.74                           & 11.62                           & 8.00 / 2.50                             & 9.76                            & 5.04                            & 22.25 \\
        SERE-K4 ($_{\rho{=}0.0}$)              & 100\%                          & $S{=}4^\ast$                     & 94.00                           & 57.08                           & 83.55                           & 29.16                           & 55.56                           & 65.27 / 50.97                           & 64.63                           & 63.93                           & 64.15 \\
        SERE-K2 ($_{\rho{=}0.1}$)                 & 100\%                          & $S{=}2^\ast$                     & 88.20                           & 49.17                           & 73.75                           & 23.20                           & 50.00                           & 16.77 / 5.24 & 42.07 & 60.28 & 50.65 \\
        \rowcolor{blue!5}
        \textbf{\sys-D64}                               & 100\%                          & 50.0\%                           & \textbf{97.00}                           & 65.31                           & 83.92                           & 28.12                           & \textbf{63.64}                           & \textbf{71.26} / \textbf{58.76}                           & 74.39                           & 66.36                           & {68.75} \\
        \rowcolor{blue!5}
        \textbf{\sys-D32}                               & 100\%                          & 25.0\%                           & 96.80                           & \textbf{67.29}                           & \textbf{84.47}                           & 28.03                           & \textbf{63.64}                           & 70.06 / 57.19                           & \textbf{75.00}                           & \textbf{66.45}                           & \textbf{68.97} \\
        \cmidrule(lr){1-12}
        \rowcolor{groupgray}\multicolumn{12}{l}{\textbf{Prefill \& Decode Acceleration}} \\
        All TopK=6 & 75\%                           & 75.0\%                           & 95.60                           & 65.31                           & 81.52                           & 27.66                           & 59.09                           & 68.86 / 57.11                           & 67.68                           & 64.60                           & 66.29 \\
        All TopK=4 & 50\%                           & 50.0\%                           & 93.60                           & 56.88                           & 74.31                           & 25.37                           & 56.31                           & 58.08 / 42.07                           & 26.83                           & 57.97                           & 56.17 \\
        {MC-MoE K=4} & {50\%} & {50.0\%} & {93.40} & {57.08} & {74.86} & {25.20} & {39.52} & {57.49 / 42.51} & {52.44} & {58.44} & {57.30} \\
        {MoDES K=4}  & {50\%} & {50.0\%} & {96.20} & {63.75} & {81.52} & {28.29} & {42.93} & {70.66 / 56.59} & \textbf{76.22} & \textbf{66.16} & {65.72} \\
        \rowcolor{blue!5}
        \textbf{\sys P4+D64}                   & 50\%                           & 50.0\%                           & \textbf{97.00}                           & \textbf{67.29}                           & \textbf{82.44}                           & \textbf{28.60}                           & \textbf{62.12}                           & \textbf{71.26 / 57.26}                           & 74.39 & 64.89 & \textbf{68.50} \\
        \rowcolor{blue!5}
        \textbf{\sys P4+D32}                   & 50\%                           & 25.0\%                           & 96.40                           & 65.83                           & 79.48                           & 28.14                           & 58.08                           & 69.46 / 56.29                           & 74.39 & 65.00 & 67.10 \\
        \bottomrule
    \end{tabular}}
    \caption{Qwen3-30B-A3B quality. \textbf{Bold} marks the best result per setting. P4 denotes 4 experts per token in prefill, D$m$ means a decode pool with size $m$, and $S^\ast$ means SERE's dynamic set. P/A.@8 means the score of Pass@8 and Avg@8.}
    \vspace{-4mm}
    \label{tab:main_qwen_results}
    
\end{table*}

\subsection{Phase-Specific Selection and Unified Folding}

At inference time, prefill and decode use phase-specific retained-expert selectors but share the same transfer criterion and folding operator, parameterized by the calibrated scalar table $\mathbf{S}^{(l)}$ and loss table $\mathbf{L}^{(l)}$. Below, we omit the layer index, denote the router weight by $w_{i,e}=\alpha_e(x_i)$, and use $h_e$ for the cached output-norm estimate of expert $e$.

\paragraph{1. Select phase-specific primary experts.}
As illustrated in Figure~\ref{fig:uniep_method}, prefill selects $K_{\mathrm{pre}}$ primary experts independently for each token, whereas decode selects a shared set of at most $D$ experts for the current batch $\mathcal{X}_q$:
\begin{align}
    B_{\mathrm{pre}}(x_i)
    &=\operatorname{TopK}_{e\in S_K(x_i)}
      \bigl(w_{i,e}h_e,K_{\mathrm{pre}}\bigr),
      \label{eq:prefill_budget}\\
    B_{\mathrm{dec}}(\mathcal{X}_q)
    &=\operatorname{TopK}_{e\in\cup_i S_K(x_i)}
      \left(\sum_{x_i\in\mathcal{X}_q}w_{i,e}h_e,D\right).
      \label{eq:decode_budget}
\end{align}
The score $w_{i,e}h_e$ estimates the magnitude of each routed contribution. Prefill applies it locally to reduce token-level computation; decode aggregates it across tokens to reduce the batch-level expert union.

\paragraph{2. Choose minimum-loss transfers.}
For either phase, each omitted source expert $E_s$ selects the retained target with the smallest calibrated reconstruction loss:
\begin{equation}
    \pi_\phi(s)=\arg\min_{t\in B_\phi}\ell_{s\to t},
    \label{eq:minimum_loss_assignment}
\end{equation}
where $B_\phi$ is the token-level set $B_{\mathrm{pre}}(x_i)$ or the batch-level set $B_{\mathrm{dec}}(\mathcal{X}_q)$. The two phases therefore share the same loss lookup and differ only in the candidate target set.

\paragraph{3. Fold router metadata.}
Let $S_{s\to t}=s_{s\to t}^{\star}$ denote the scalar-table lookup. In prefill, omitted routes mapped to the same target are coalesced by updating the target weight:
\begin{equation}
    \widetilde{w}_{i,t}
    =
    w_{i,t}
    +
    \sum_{\substack{s\in S_K(x_i)\setminus B_{\mathrm{pre}}(x_i)\\
                    \pi_{\mathrm{pre}}(s)=t}}
    w_{i,s}S_{s\to t},
    \quad t\in B_{\mathrm{pre}}(x_i).
    \label{eq:prefill_fold}
\end{equation}
The fused MoE kernel consequently executes only $K_{\mathrm{pre}}$ routes. In decode, each omitted route is instead remapped in place:
\begin{equation}
    (E_s,w_{i,s})
    \longmapsto
    \left(E_{\pi_{\mathrm{dec}}(s)},
    w_{i,s}S_{s\to\pi_{\mathrm{dec}}(s)}\right).
    \label{eq:decode_remap}
\end{equation}
Retained routes remain unchanged, while every remapped expert belongs to $B_{\mathrm{dec}}(\mathcal{X}_q)$; hence the batch touches at most $D$ distinct experts without changing the regular Top-$K$ routing layout. Both transformations modify only router metadata before the existing fused MoE kernel.

\section{Experiments}

\begin{table}[t]
    \centering
    \setlength{\tabcolsep}{2.5pt}
    \resizebox{\columnwidth}{!}{%
        \begin{tabular}{l|ccccc|c}
            \toprule
            \rowcolor{gray!30}
            \textbf{Method} & \textbf{AIME24} & \textbf{IFEval} & \textbf{GPQA} & \textbf{Eval+} & \textbf{MMLU$_{pro}$} & \textbf{Avg.} \\
            \midrule
            \rowcolor{groupgray}\multicolumn{7}{l}{\textbf{Baseline}} \\
            Original Top-8 & 48.75 & 83.55 & 76.26 & 71.34 & 58.98 & 67.78 \\
            \cmidrule(lr){1-7}
            \rowcolor{groupgray}\multicolumn{7}{l}{\textbf{Prefill-Only Acceleration}} \\
            Prefill TopK=6 & 46.67 & \textbf{83.55} & \textbf{75.25} & 73.78 & 56.53 & \textbf{67.16} \\
            Prefill TopK=4 & 45.00 & 80.59 & 69.19 & \textbf{75.61} & 40.41 & 62.16 \\
            \rowcolor{blue!5}
            \textbf{\sys-P4} & \textbf{48.75} & 80.78 & 73.74 & 67.68 & \textbf{58.57} & 65.90 \\
            \cmidrule(lr){1-7}
            \rowcolor{groupgray}\multicolumn{7}{l}{\textbf{Decode-Only Acceleration}} \\
            REAP-D64 & 40.42 & 69.13 & 69.70 & 42.07 & 55.92 & 55.45 \\
            \rowcolor{blue!5}
            \textbf{\sys-D64} & \textbf{47.50} & \textbf{83.73} & \textbf{71.21} & \textbf{73.17} & \textbf{57.96} & \textbf{66.71} \\
            \cmidrule(lr){1-7}
            \rowcolor{groupgray}\multicolumn{7}{l}{\textbf{Prefill \& Decode Acceleration}} \\
            All TopK=6 & 46.67 & 80.96 & 72.22 & 66.46 & 41.43 & 61.55 \\
            All TopK=4 & 36.67 & 75.60 & 71.21 & \textbf{70.73} & 38.78 & 58.60 \\
            \rowcolor{blue!5}
            \textbf{\sys P4+D64} & 44.17 & 80.22 & \textbf{72.73} & 67.68 & \textbf{58.98} & 64.76 \\
            \rowcolor{blue!5}
            \textbf{\sys P4+D32} & \textbf{50.00} & \textbf{81.70} & 71.21 & 67.68 & 57.14 & \textbf{65.55} \\
            \bottomrule
        \end{tabular}%
    }
    \caption{GLM-4.5-Air quality. \textbf{Bold} for the best per setting.}
    \vspace{-5mm}
    \label{tab:glm45air_results}
\end{table}

\begin{figure*}[!htp]
    \centering
    \includegraphics[width=.48\textwidth]{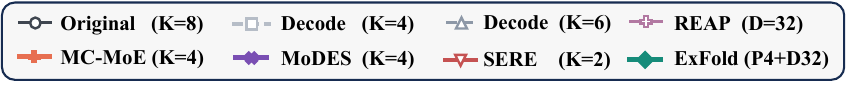}
    \par\vspace{0mm}
    \begin{minipage}[t]{.37\textwidth}
        \centering
        \includegraphics[width=\linewidth]{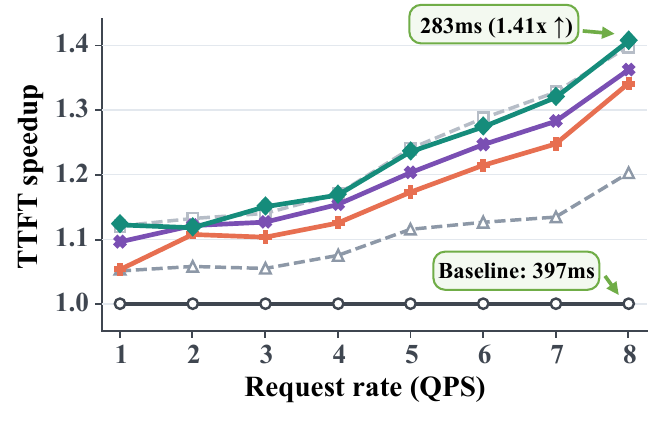}
        \vspace{-5mm}
        \par\small\textbf{(a)} Prefill acceleration.
    \end{minipage}\hfill
    \begin{minipage}[t]{.35\textwidth}
        \centering
        \includegraphics[width=\linewidth]{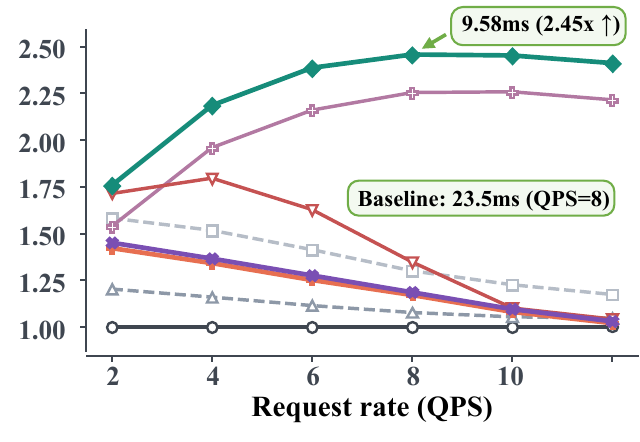}
        \vspace{-5mm}
        \par\small\textbf{(b)} Decode acceleration
    \end{minipage}\hfill
    \begin{minipage}[t]{.21\textwidth}
        \centering
        \includegraphics[width=\linewidth]{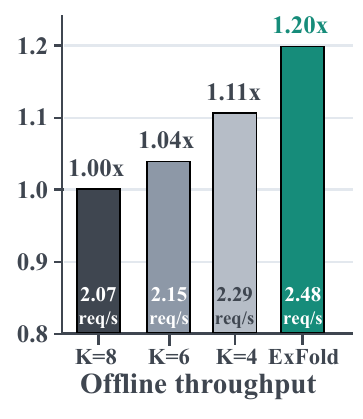}
        \vspace{-5mm}
        \par\small\textbf{(c)} Offline acceleration
    \end{minipage}
    \caption{Online TTFT/TPOT speedups and offline serving throughput.}
    \label{fig:speed_prefill}
\end{figure*}

\begin{table*}[!ht]
    \centering
    \setlength{\tabcolsep}{2.6pt}
    \resizebox{.96\textwidth}{!}{%
        \begin{tabular}{l|cccccccc|l|l|l}
            \toprule
            \rowcolor{gray!30}
            \textbf{Method}   & \textbf{MATH$_{500}$} & \textbf{AIME24} & \textbf{IFEval} & \textbf{IFBench} & \textbf{GPQA}  & \textbf{Eval+} & \textbf{MMLU$_{pro}$} & \textbf{PPL$\downarrow$}   & \textbf{$\Delta$Storage} & \textbf{$\Delta$Token Cost} & \textbf{Kernel} \\
            \midrule
            Original TopK=8          & 97.40   & 64.48  & 83.73  & 29.31   & 63.13 & 77.44 & 68.57        & {3.020} & 0          & $O(4LC_{\mathrm{FFN}})$ & FusedMoE \\
            \midrule
            Prefill TopK=4            & 96.00   & 66.56  & 76.00  & 29.02   & 60.61 & 73.17 & 65.28        & {3.470} & 0          & 0                      & FusedMoE \\
            Global scalar           & 96.00   & 65.21  & 80.96  & 25.73   & 59.60 & 75.00 & 65.60        & {3.373} & 4 B        & $O(4L)$                          & FusedMoE                \\
            Layer scalar            & 96.60   & \textbf{67.60}  & 82.44  & 26.30   & 61.11 & 74.39 & 64.70        & {3.350} & 192 B      & $O(4L)$     & FusedMoE                \\
            \rowcolor{blue!5}
            \textbf{Expert scalar} & \textbf{97.00}   & 66.46  & \textbf{84.00}  & \textbf{29.53}   & \textbf{62.63} & \textbf{76.83} & 66.94        & {3.352} & 6.00 MiB   & $O(4L)$  & FusedMoE \\
            Diagonal                & 96.60 & 67.50 & 83.92 & 27.63 & 60.61 & 72.56 & 65.87 & \textbf{3.308} & 316.7 MiB & $O(4LH)$    & Unfused \\
	            Low-rank R8             & 95.60 & 65.00 & 83.92 & 27.35 & 60.10 & 74.39 & \textbf{67.35} & {3.341} & 2.44 GiB   & $O(64LH)$  & Unfused \\
            Low-rank R16            & 95.20   & 62.50 & 83.55 & 27.65 & 61.62 & 75.61 & 66.60 & {3.342} & 4.87 GiB   & $O(128LH)$  & Unfused \\
            \bottomrule
        \end{tabular}
    }
    \caption{Projector ablation under Top-4 prefill. Overheads use FP32 state and are relative to Prefill TopK=4. Here $\Delta K=4$; R8/R16 incur $2\Delta KR=64/128$ operations per hidden dimension. $L,H,C_{\mathrm{FFN}}$: layers, hidden size, and one expert-call time.}
    \vspace{-3mm}
    \label{tab:prefill_quick_results}
\end{table*}

\subsection{Experimental Setup}

\paragraph{Models.}
We evaluate Qwen3-30B-A3B as the primary model and GLM-4.5-Air, DeepSeek-V2-Lite, DeepSeek-V4-Flash, Qwen3.5-35B-A3B (in Appendix C) for cross-architecture generalization.


\paragraph{Benchmarks.}
We evaluate mathematical reasoning with MATH500 and AIME24, code generation with LiveCodeBenchV5 and HumanEval+, instruction following with IFEval and IFBench, and knowledge reasoning with GPQA-Diamond and MMLU-Pro~\cite{hendrycks2021math,jain2024livecodebench,liu2023evalplus,zhou2023ifeval,ifbench2025,rein2023gpqa,wang2024mmlupro}. For efficiency, we report TTFT under prefill-dominated serving, TPOT under generation-dominated serving, and offline throughput in vLLM~\cite{kwon2023vllm}. Metric and workload details are deferred to Appendix A.

\paragraph{Baselines.}
For prefill, we compare with direct Top-$K$ reduction, MC-MoE, and MoDES under matched token-wise expert budgets. For decode, we compare with static expert pruning (REAP) and dynamic expert skipping (SERE). We report prefill-only, decode-only, and joint acceleration to distinguish phase-specific quality loss from errors accumulated across both phases~\cite{huang2024harder,lasby2025reap,wu2026sere}.

\begin{figure*}[!ht]
    \centering
    \includegraphics[width=.96\textwidth]{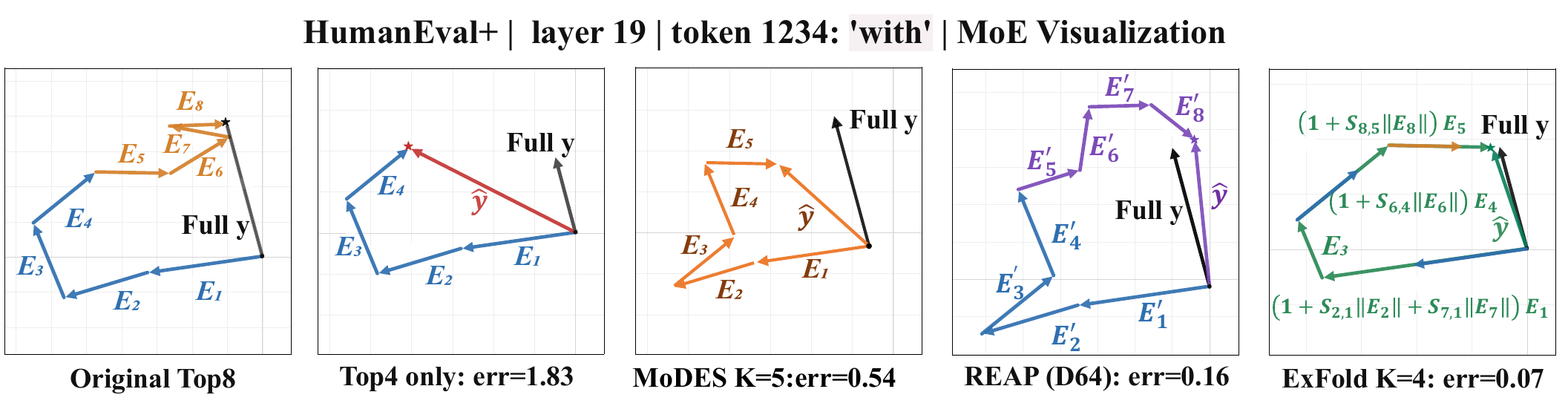}
    \vspace{-2mm}
    \caption{Per-token MoE output reconstruction under expert reduction. Error denotes $\|\widehat{y}-y\|_2$.}
    \label{fig:folding_case_study}
    \vspace{-3mm}
\end{figure*}

\paragraph{Hyperparameters.}
We calibrate \sys on 32 unlabeled sequences of at most 4096 tokens, use source-output-norm weighting, and set the ridge coefficient to $\lambda=10^{-3}$. $\mathrm{P}X$ denotes $X$ retained experts per prefill token, while $\mathrm{D}X$ denotes at most $X$ active experts per decode batch. We implement \sys with a custom Triton operator in vLLM and run efficiency experiments in BF16 on NVIDIA H800 GPUs~\cite{tillet2019triton}; complete calibration, hardware, and serving configurations are provided in Appendix A.

\subsection{Main Quality Results}

Tables~\ref{tab:main_qwen_results} and~\ref{tab:glm45air_results} compare \sys with phase-specific baselines under matched execution budgets. We focus on whether quality is preserved, rather than treating a lower expert count alone as an improvement.

\paragraph{Prefill-only acceleration.}
At the P4 budget, \sys preserves the original Qwen3 average and outperforms all compute-matched baselines (69.26 versus 66.64 for Direct Top-4). The gains are most pronounced on instruction following and code generation, where hard dropping loses important routed contributions. Thus, folding can halve prefill expert computation without the quality loss of direct sparsification.

\paragraph{Decode-only acceleration.}
The advantage of \sys widens as the batch-level expert budget becomes tighter. At D32, it remains within 0.07 points of the original model; SERE-K2 recovers part of the quality lost by static pruning but remains substantially lower. This trend shows that recovering omitted expert contributions becomes increasingly important when the active expert pool is aggressively constrained.

\paragraph{Joint prefill and decode acceleration.}
Applying Direct Top-4 to both phases compounds approximation error and reduces the average to 56.17, while \sys P4+D64 retains 68.50. \sys also remains stronger than MoDES when both phases are accelerated. The shared projector therefore avoids stacking two independent dropping errors and supports both phase-specific budgets under one approximation objective.

\paragraph{Cross-architecture generalization.}
On GLM-4.5-Air, \sys retains 97.2\% and 98.4\% of the original average quality under prefill-only P4 and decode-only D64, respectively. When both phases are compressed, P4+D32 still retains 96.7\%. These results confirm that \sys preserves quality beyond the primary model.
\textcolor{candidateRed}{Additional DeepSeek-V2-Lite and Qwen3.5 results can be seen in Appendix C.}

\begin{table}[!ht]
    \centering
    \small
    \setlength{\tabcolsep}{3.2pt}
    \resizebox{\columnwidth}{!}{%
    \begin{tabular}{lccccc}
        \toprule
        \rowcolor{gray!30}
        \textbf{Calibration} & \textbf{MATH$_{500}$} & \textbf{GPQA}  & \textbf{LCB (P/A.@8)}  & \textbf{Eval+} & \textbf{Avg.} \\
        \midrule
        GPQA             & 96.40 & \textbf{65.15} & 70.06 / 56.96 & 71.95 & 72.10 \\
        MATH             & 96.40 & 60.10 & 69.46 / 56.36 & 73.17 & 71.10 \\
        CODE             & 96.60 & 59.09 & 70.06 / 56.74 & 75.61 & 71.62 \\
        PRETRAIN         & \textbf{97.00} & 63.13 & 69.46 / 57.34 & \textbf{76.83} & 72.75 \\
        \rowcolor{blue!5}
        MIXED            & \textbf{97.00} & 62.63 & \textbf{70.66 / 57.86} & \textbf{76.83} & \textbf{73.00} \\
        \bottomrule
    \end{tabular}
    }
    \vspace{-1mm}
    \caption{Calibration-corpus ablation for \sys-P4.}
    \vspace{-2mm}
    \label{tab:calib_data_ablation}
\end{table}

\begin{table}[!ht]
    \centering
    \small
    \setlength{\tabcolsep}{2.8pt}
    \resizebox{\columnwidth}{!}{%
    \begin{tabular}{lccccc}
        \toprule
        \rowcolor{gray!30}
        \textbf{Selection criterion} & \textbf{MATH$_{500}$} & \textbf{IFEval} & \textbf{GPQA} & \textbf{Eval+} & \textbf{Avg.} \\
        \midrule
        \rowcolor{groupgray}\multicolumn{6}{l}{\textbf{Prefill (\sys-P4)}} \\
        rank by $S_i$ & \textbf{97.00} & 82.62 & \textbf{64.65} & 73.17 & 79.36 \\
        rank by $H_i$ & \textbf{97.00} & 67.47 & 57.58 & 75.61 & 74.42 \\
        \rowcolor{blue!5}
        \textbf{rank by $S_i\times H_i$ (Ours)} & \textbf{97.00} & \textbf{84.00} & 62.63 & \textbf{76.83} & \textbf{80.11} \\
        \cmidrule(lr){1-6}
        \rowcolor{groupgray}\multicolumn{6}{l}{\textbf{Decode (\sys-D64)}} \\
        rank by $S_i$ & \textbf{97.00} & 83.92 & 60.10 & 73.17 & 78.55 \\
        rank by $H_i$ & 96.40 & \textbf{84.10} & 63.13 & 71.34 & 78.74 \\
        \rowcolor{blue!5}
        \textbf{rank by $S_i\times H_i$ (Ours)} & \textbf{97.00} & 83.92 & \textbf{63.64} & \textbf{74.39} & \textbf{79.74} \\
        \bottomrule
    \end{tabular}
    }
    \caption{Retained-expert selection on Qwen3-30B-A3B.}
    \vspace{-5mm}
    \label{tab:retained_expert_selection}
\end{table}

\subsection{Efficiency Evaluation}

\paragraph{Online prefill latency.}
Figure~\ref{fig:speed_prefill}(a) shows that reducing token-level expert computation translates directly into lower TTFT. \sys reaches a $1.41\times$ speedup at 8 QPS and closely tracks compute-matched Top-4 methods, indicating little overhead from scalar folding. Unlike direct reduction, it realizes this compute benefit while preserving model quality.

\paragraph{Online decode latency.}
Figure~\ref{fig:speed_prefill}(b) reveals a different scaling trend in decode: token-wise Top-4 and Top-6 lose their benefit as QPS increases because the batch activates a broader expert union. By bounding this union, \sys reaches a $2.45\times$ TPOT speedup at 8 QPS and sustains about $2.4\times$ through 12 QPS. This confirms that reducing batch-level weight traffic, rather than token-wise FLOPs alone, is essential for efficient decode.

\paragraph{Offline serving throughput.}
Figure~\ref{fig:speed_prefill}(c) shows that the online gains transfer to throughput-oriented serving: \sys improves offline throughput by $1.20\times$ and exceeds Direct Top-4. This result also confirms that the Triton folding operator adds little runtime overhead under large batches.

\subsection{Scaling to DeepSeek-V4-Flash}
\label{sec:deepseek_v4_extension}

DeepSeek-V4-Flash is a 284B-parameter MoE with 256 routed experts and Top-6
routing~\cite{deepseekv4}. It provides a substantially larger expert space than
the models above and therefore tests whether folding remains effective when
both the number of candidate experts and the batch-level expert union grow.
We use P3 to retain three routed experts per prefill token and D128/D64 to cap
the decode expert pool at 50\%/25\% of the routed experts. All quality rows use
the same full-suite evaluator protocol; Appendix~\ref{sec:dsv4_repro} reports
the task sizes and sampling rules.

\begin{table*}[!t]
    \centering
    \small
    \setlength{\tabcolsep}{3pt}
    \resizebox{.96\textwidth}{!}{%
    \begin{tabular}{l|cc|ccccccccc|c}
        \toprule
        \rowcolor{gray!30}
        \textbf{Method} & \textbf{Prefill} & \textbf{Decode} &
        \textbf{MATH$_{500}$} & \textbf{AIME25} &
        \textbf{AIME26} & \textbf{IFEval} & \textbf{IFBench} &
        \textbf{GPQA} & \textbf{LCB} & \textbf{Eval+} &
        \textbf{MMLU$_{pro}$} & \textbf{Avg.} \\
        \midrule
        \rowcolor{groupgray}\multicolumn{13}{l}{\textbf{Baseline}} \\
        Original TopK=6
            & 100\% & 100\% & 93.40 & 70.42 & 72.50 & 82.44 & 36.23
            & 73.23 & 54.90 & 89.25 & 81.76 & 72.68 \\
        \cmidrule(lr){1-13}
        \rowcolor{groupgray}\multicolumn{13}{l}{\textbf{Prefill-Only Acceleration}} \\
        Prefill TopK=3
            & 50\% & 100\% & 92.00 & \textbf{69.17} & 69.17 & 80.22 & \textbf{35.67}
            & 69.70 & 54.12 & 85.59 & 79.20 & 70.54 \\
        \rowcolor{blue!5}
        \textbf{\sys-P3}
            & 50\% & 100\% & \textbf{93.20} & 65.83 & \textbf{70.42} & \textbf{82.99} & 35.28
            & \textbf{72.22} & \textbf{55.67} & \textbf{87.65} & \textbf{80.48}
            & \textbf{71.53} \\
        \cmidrule(lr){1-13}
        \rowcolor{groupgray}\multicolumn{13}{l}{\textbf{Decode-Only Acceleration}} \\
        Decode TopK=3
            & 100\% & 50\% & 88.80 & 47.08 & 49.17 & 79.48 & 32.01
            & 73.23 & 49.74 & 82.09 & 80.68 & 64.70 \\
        REAP-D128
            & 100\% & 50\% & 93.20 & 70.00 & \textbf{74.58} & 71.16 & 34.30
            & 72.73 & 52.58 & \textbf{89.86} & 80.73 & 71.02 \\
        \rowcolor{blue!5}
        \textbf{\sys-D128}
            & 100\% & 50\% & 94.20 & \textbf{70.83} & 72.50 & 81.15 & 35.67
            & \textbf{75.76} & \textbf{54.12} & 88.41 & \textbf{81.82}
            & \textbf{72.72} \\
        \rowcolor{blue!5}
        \textbf{\sys-D64}
            & 100\% & 25\% & \textbf{94.80} & \textbf{70.83} & 71.25 & \textbf{81.33} & \textbf{37.52}
            & 72.73 & 53.87 & 88.72 & 81.79
            & 72.54 \\
        \cmidrule(lr){1-13}
        \rowcolor{groupgray}\multicolumn{13}{l}{\textbf{Prefill \& Decode Acceleration}} \\
        All TopK=3
            & 50\% & 50\% & 87.40 & 52.08 & 51.25 & 75.97 & 28.10
            & 68.18 & 47.94 & 85.98 & 79.85 & 64.08 \\
        \rowcolor{blue!5}
        \textbf{\sys P3+D128}
            & 50\% & 50\% & \textbf{92.80} & \textbf{72.50} & \textbf{72.50} & \textbf{81.52} & \textbf{35.45}
            & 68.69 & \textbf{55.41} & \textbf{90.17} & \textbf{80.61}
            & \textbf{72.18} \\
        \rowcolor{blue!5}
        \textbf{\sys P3+D64}
            & 50\% & 25\% & 92.60 & 66.67 & 70.83 & 80.96 & 31.56
            & \textbf{74.75} & 54.38 & 86.97 & 80.21
            & 70.99 \\
        \bottomrule
    \end{tabular}}
    \caption{DeepSeek-V4-Flash quality. \textbf{Bold} marks the best result per setting. P3 denotes 3 experts per token in prefill, and D$m$ means a decode pool with size $m$.}
    \label{tab:dsv4_quality}
    \vspace{1mm}
    \includegraphics[width=.96\textwidth]{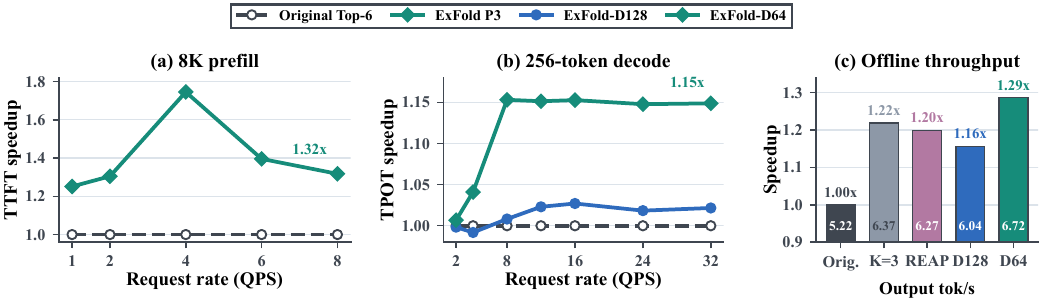}
    \captionsetup{type=figure}
    \vspace{-3mm}
    \caption{DeepSeek-V4-Flash speedups on H800: \textbf{(a)} 8K TTFT,
    \textbf{(b)} 256-token TPOT, and \textbf{(c)} offline throughput.}
    \vspace{-4mm}
    \label{fig:dsv4_speed}
\end{table*}

Table~\ref{tab:dsv4_quality} shows that direct all-stage K=3 reduction retains
only 88.17\% of the original average, with the largest losses on AIME25/26.
\sys P3+D64 instead retains 97.68\%, a 9.51-point retention gain under the
same three-expert prefill budget, while P3+D128 retains 99.32\%. Decode-only
D64 and D128 remain within 0.2 average points of Original. These results
support a quality frontier rather than one universal budget: D128 is the
near-lossless setting, whereas D64 trades 1.64 additional quality-retention
points for a tighter expert pool.

Figure~\ref{fig:dsv4_speed} follows the same online/offline structure as
Figure~\ref{fig:speed_prefill}. P3 improves mean 8K TTFT across QPS 1--8,
reaching 1.32$\times$ at QPS 8. In decode, D64 stabilizes near
1.15$\times$ TPOT speedup once the batch is saturated, while D128 remains the
quality-first operating point. Under the saturated throughput workload, D64
raises output throughput from 5.22k to 6.72k tokens/s (1.29$\times$), ahead
of Direct K=3, REAP-D128, and D128. The code release exposes both budgets as
serving arguments and includes the calibrated matrix used by these runs.

\subsection{Ablations and Analysis}
\label{sec:ablation_analysis}

\paragraph{Which projector parameterization is practical?}
Table~\ref{tab:prefill_quick_results} compares projector families under the same Top-4 budget. Pairwise scalars outperform global and layer-wise scaling, while diagonal and low-rank projectors require substantially more state and unfused vector transforms. Although the diagonal projector lowers perplexity, the pairwise scalar performs better on six of seven downstream metrics; low-rank variants further add 2.44--4.87 GiB of state without consistent gains. \underline{{Pairwise scalars give the best trade-off.}}
\textcolor{candidateRed}{In appendix B, we detail each implementation.}

\paragraph{Does \sys require task-specific calibration?}
Table~\ref{tab:calib_data_ablation} varies only the calibration domain. Mixed calibration achieves the best average (73.00), but all domains remain within 1.9 points and task-matched data is not consistently optimal. This stability suggests that calibration captures expert geometry rather than benchmark-specific behavior. \underline{{No task-specific calibration is required.}}

\paragraph{How should \sys select retained experts?}
Table~\ref{tab:retained_expert_selection} compares router score $S_i$, output magnitude $H_i$, and their product under fixed budgets. Router score omits output scale, while magnitude alone omits routing relevance; their product approximates contribution magnitude and yields the best average in both phases. \underline{Rank retained experts by $S_i\times H_i$.}

\paragraph{Why does expert folding preserve quality?}
Figure~\ref{fig:folding_case_study} visualizes the routed expert contributions and resulting MoE output for a representative token. Direct Top-4 and MoDES omit routed vectors and produce large reconstruction errors of 1.83 and 0.54, respectively. REAP reduces the global expert pool and lowers the error to 0.16 by selecting Top-8 within the retained pool, but it still executes eight experts per token and therefore does not reduce expert FLOPs~\cite{huang2025modes,lasby2025reap}. ExFold instead transfers omitted contributions to directionally aligned retained experts and folds their calibrated scales into the gate weights, achieving the smallest error of 0.07 with only four executed experts. \underline{{Folding recovers omitted contributions.}}
\textcolor{candidateRed}{Additional token-level cases are shown in Appendix E.}
\FloatBarrier

    \enlargethispage{3\baselineskip}
    \section{Conclusion}
\enlargethispage{2\baselineskip}

We presented \sys, a training-free framework for jointly accelerating MoE prefill and decode. It combines phase-specific retained-set selectors with a shared directed projector that folds excluded contributions into router metadata. Across MoE architectures, \sys improves TTFT, TPOT, and serving throughput while preserving more quality than expert dropping. Because folding changes only router metadata, it remains complementary to kernel, scheduling, and parallelism optimizations. These results establish output recovery as a practical basis for unified MoE inference acceleration.

\FloatBarrier
\newpage

    \bibliography{aaai2027}

    \clearpage

    \appendix
    \section{Reproducibility Details}
\label{sec:appendix}
\label{sec:appendix_setup}

\paragraph{Model architectures.}
Table~\ref{tab:appendix_model_architectures} summarizes the evaluated released checkpoints. We report the architecture exposed by each checkpoint configuration; shared experts are executed in addition to the routed Top-$K$ experts. ``Expert FFN'' is the hidden width of each expert feed-forward network.

\begin{table*}[t]
    \centering
    \small
    \setlength{\tabcolsep}{5.0pt}
    \resizebox{0.92\textwidth}{!}{%
    \begin{tabular}{lccccccc}
        \toprule
        \rowcolor{gray!30}
        \textbf{Model} & \textbf{Layers} & \textbf{MoE Layers} &
        \textbf{Routed Experts} & \textbf{Shared Experts} &
        \textbf{Top-$K$} & \textbf{Hidden Size} & \textbf{Expert FFN} \\
        \midrule
        Qwen3-30B-A3B~\cite{qwen3a3b} & 48 & 48 & 128 & 0 & 8 & 2048 & 768 \\
        GLM-4.5-Air~\cite{glm45air} & 46 & 45 & 128 & 1 & 8 & 4096 & 1408 \\
        DeepSeek-V2-Lite-Chat~\cite{deepseekv2} & 27 & 26 & 64 & 2 & 6 & 2048 & 1408 \\
        Qwen3.5-35B-A3B~\cite{qwen35a3b} & 40 & 40 & 256 & 1 & 8 & 2048 & 512 \\
        \bottomrule
    \end{tabular}}
    \caption{Architectures of the evaluated MoE checkpoints.}
    \label{tab:appendix_model_architectures}
\end{table*}

\paragraph{Benchmarks and metrics.}
Table~\ref{tab:appendix_benchmarks} lists the evaluated benchmarks. ``Cases'' is the number of problems in the evaluated benchmark snapshot, and ``Samples'' is the number of generations per problem. The GLM-4.5-Air MMLU-Pro comparison uses a fixed 490-case subset containing the first 35 examples from each of 14 categories. For IFBench, we evaluate the 294-case OpenCompass snapshot and average prompt- and instruction-level accuracy under strict and loose checking. For LiveCodeBench v5, we use the same fixed 167-problem subset for every method and generate eight solutions per problem. The Qwen3-30B-A3B evaluation uses 32 samples for each AIME 2024 problem, whereas the GLM-4.5-Air evaluation uses eight. LiveCodeBench reports pass@8 and mean correctness over the eight samples; pass@8 is its primary score in aggregate comparisons.

\begin{table*}[t]
    \centering
    \small
    \setlength{\tabcolsep}{5.5pt}
    \resizebox{0.90\textwidth}{!}{%
    \begin{tabular}{llccl}
        \toprule
        \rowcolor{gray!30}
        \textbf{Benchmark} & \textbf{Domain} & \textbf{Cases} &
        \textbf{Samples} & \textbf{Metric} \\
        \midrule
        MATH500~\cite{hendrycks2021math,math500snapshot} & Mathematics & 500 & 1 & Symbolic-equivalence accuracy \\
        AIME24~\cite{aime2024} & Competition mathematics & 30 & 32 (Qwen3); 8 (GLM) & Mean symbolic-equivalence accuracy \\
        IFEval~\cite{zhou2023ifeval} & Instruction following & 541 & 1 & Prompt-level strict accuracy \\
        IFBench~\cite{ifbench2025} & Instruction following & 294 & 1 & OpenCompass four-way mean \\
        GPQA-Diamond~\cite{rein2023gpqa} & Graduate-level reasoning & 198 & 1 & Accuracy \\
        LiveCodeBench v5~\cite{jain2024livecodebench} & Code generation & 167 & 8 & Pass@8 / mean@8 \\
        HumanEval+~\cite{liu2023evalplus} & Code generation & 164 & 1 & Pass@1 \\
        MMLU-Pro~\cite{wang2024mmlupro} & Knowledge reasoning & 12,032 (490 for GLM) & 1 & Category-macro accuracy \\
        \bottomrule
    \end{tabular}}
    \caption{Benchmarks, evaluation sizes, and reported metrics.}
    \label{tab:appendix_benchmarks}
\end{table*}

\paragraph{Calibration protocol.}
For the primary Qwen3 configuration, we calibrate once on 32 unlabeled sequences of at most 4,096 tokens and keep the resulting projector fixed across downstream tasks. DeepSeek-V2-Lite uses the same calibration size and token limit; Qwen3.5 aggregates four disjoint eight-sequence calibration shards. GLM-4.5-Air uses a separate 16-sequence capture with at most 1,024 tokens per sequence. The 32-sequence pool is drawn from the chat, code, and science, technology, engineering, and mathematics (STEM) portions of the public NVIDIA Nemotron Post-Training Dataset~\cite{nemotronposttrain}. All calibrations accumulate only forward activations, without task labels, gradients, or evaluator feedback. For each ordered co-routed expert pair, we accumulate the sufficient statistics in Eq.~\ref{eq:appendix_shared_scalar} with source-output-norm weighting and $\lambda=10^{-3}$. Scalar coefficients are clipped to $[-4,4]$, self-pair coefficients are fixed to one, and unobserved pairs receive loss $10^{30}$ so that an observed target is preferred whenever one exists.

\paragraph{Quality evaluation.}
Generation is orchestrated by an OpenCompass-compatible pipeline~\cite{opencompass2026}. IFEval disables thinking and uses prompt-level strict scoring, while IFBench uses the four-way composite defined above. LiveCodeBench and HumanEval+ are scored by executing generated programs against their benchmark test suites with the LiveCodeBench and EvalPlus evaluators~\cite{jain2024livecodebench,liu2023evalplus}. Within each model--benchmark comparison, prompts, decoding parameters, sample counts, and scoring code are fixed. Repeated generations are scored individually before aggregation. All reported scores are percentages, and ``Avg.'' is the arithmetic mean of the displayed primary metrics.

\paragraph{System and serving configuration.}
Efficiency experiments run on eight-GPU NVIDIA H800 80GB servers. The primary H800 runtime uses Ubuntu 24.04.2 LTS, NVIDIA driver 550.163.01, vLLM 0.10.2~\cite{kwon2023vllm}, PyTorch 2.8.0, CUDA 12.8, and Triton 3.4.0~\cite{tillet2019triton}. Prefill serving uses tensor parallelism four, sweeps 1--8 queries per second (QPS) with 8,192-token prompts and one generated token, and reports mean time to first token (TTFT). The matched decode comparison uses vLLM 0.11.0 and tensor parallelism one; it sweeps 2--12 QPS over 512 requests with one input token and 256 generated tokens and reports mean time per output token (TPOT). Offline serving uses tensor parallelism eight and reports completed-request throughput for 512 requests with 2,048-token inputs and 512 generated tokens. ExFold speedups are normalized to an Original run with the same hardware, engine configuration, requests, and decoding parameters. DeepSeek-V2-Lite and Qwen3.5 quality generation use bfloat16 (BF16) and tensor parallelism one on H800 or H20 GPUs; GLM-4.5-Air uses tensor parallelism four on H800 GPUs.

\section{Projector Parameterizations and Implementation}
\label{sec:appendix_projectors}

Let $L$, $E$, and $H$ denote the number of MoE layers, routed experts per
layer, and hidden dimensions. For a calibration token $x_i$ that co-routes a
source expert $E_s$ and target expert $E_t$, we write
$\mathbf{u}_i=E_s(x_i)$ and $\mathbf{v}_i=E_t(x_i)$; $l$, $s$, and $t$ index
the layer, source expert, and target expert.
For the projector-family ablation, every family uses the same fixed
calibration sequences, source-output-norm weighting, and prefill Top-4 budget.
Each sharing pattern computes its own reconstruction losses for target
assignment. The diagonal and low-rank alternatives execute through a custom
unfused reconstruction path. Table~\ref{tab:appendix_projector_implementations}
counts transformation parameters and omits the per-pair assignment-loss
table. Let $P\leq LE^2$ denote the number of materialized directed
source--target pairs.

\paragraph{Scalar sharing patterns.}
Let $\mathcal{G}$ denote a group of calibration tuples $(l,s,t,i)$. A scalar shared by that group is fitted by
\begin{equation}
    s_{\mathcal{G}}^{\star}
    =
    \frac{
        \sum_{(l,s,t,i)\in\mathcal{G}}
        w_i\,\mathbf{u}_i^{\top}\mathbf{v}_i
    }{
        \sum_{(l,s,t,i)\in\mathcal{G}}
        w_i\,\|\mathbf{v}_i\|_2^2+\lambda
    }.
    \label{eq:appendix_shared_scalar}
\end{equation}
Here $w_i=\|\mathbf{u}_i\|_2$ is the source-output-norm weight used by the calibration implementation. The global variant uses one group for the entire model, the layer-wise variant uses one group per MoE layer, and the pairwise variant uses one group for each directed source--target pair in each layer. Only the pairwise form preserves directed expert-level differences while remaining a scalar metadata update.
For target assignment, we use the normalized reconstruction loss
\begin{equation}
    \ell_{s\to t}
    =
    \frac{
        \sum_i w_i
        \|\mathbf{u}_i-s_{s\to t}^{\star}\mathbf{v}_i\|_2^2
    }{
        \sum_i w_i\|\mathbf{u}_i\|_2^2+\epsilon
    },
    \label{eq:appendix_projection_loss}
\end{equation}
where $\epsilon>0$ prevents division by zero. Lower loss indicates that
$E_t$ better reconstructs the contribution of $E_s$.

\paragraph{Diagonal projector.}
The diagonal alternative fits one coefficient per hidden dimension. Its
element-wise weighted ridge solution is
\begin{equation}
    \mathbf{d}_{s\to t}^{\star}
    =
    \frac{\sum_i w_i(\mathbf{v}_i\odot\mathbf{u}_i)}
         {\sum_i w_i(\mathbf{v}_i\odot\mathbf{v}_i)
          +\lambda\mathbf{1}},
    \label{eq:appendix_diagonal_projector}
\end{equation}
where the division is element-wise.
The resulting $\operatorname{Diag}(\mathbf{d}_{s\to t}^{\star})$ cannot be
absorbed into a router weight and must act on the hidden vector online.

\paragraph{Scalar-plus-low-rank projector.}
Let $\mathbf{U},\mathbf{V}\in\mathbb{R}^{m\times H}$ stack source and target outputs, and $\mathbf{W}=\operatorname{Diag}(w_1,\ldots,w_m)$. We fit a layer-wise orthonormal basis $\mathbf{Q}^{(l)}\in\mathbb{R}^{H\times R}$ to centered scalar-residual samples by randomized truncated singular value decomposition (SVD) and parameterize
\begin{equation}
    \mathbf{M}_{s\to t}
    =
    s_{s\to t}^{\star}\mathbf{I}
    +
    \mathbf{Q}^{(l)}\mathbf{C}_{s\to t},
    \qquad
    \mathbf{C}_{s\to t}\in\mathbb{R}^{R\times H}.
    \label{eq:appendix_lowrank_parameterization}
\end{equation}
Defining $\mathbf{Z}=\mathbf{W}^{1/2}\mathbf{V}\mathbf{Q}^{(l)}$,
$\mathbf{R}_{s\to t}=\mathbf{W}^{1/2}
(\mathbf{U}-s_{s\to t}^{\star}\mathbf{V})$, and
$\mathbf{G}=\mathbf{Z}^{\top}\mathbf{Z}$, weighted ridge regression gives
\begin{equation}
    \mathbf{C}_{s\to t}^{\star}
    =
    (\mathbf{G}+\lambda\mathbf{I})^{-1}
    \mathbf{Z}^{\top}\mathbf{R}_{s\to t}.
    \label{eq:appendix_lowrank_solution}
\end{equation}
The evaluated artifact collects a rank-32 basis and retains at most 1,024
directed pairs per layer ranked by
accumulated source-weighted source energy. We evaluate its first
$R\in\{8,16\}$ directions; pairs outside the materialized support use only
the scalar term. The basis and pair-specific correction require an online
vector transform and substantial state, preventing reuse of the standard
fused MoE path.

\begin{table}[!ht]
    \centering
    \footnotesize
    \setlength{\tabcolsep}{1.8pt}
    \begin{tabular}{@{}lccc@{}}
        \toprule
        \rowcolor{gray!30}
        \textbf{Projector} & \textbf{Storage} &
        \textbf{Online} & \textbf{Kernel} \\
        \midrule
        Global scalar & $1$ & Scalar & Fused \\
        Layer scalar & $L$ & Scalar & Fused \\
        Pair scalar & $P$ & Scalar & Fused \\
        Pair diagonal & $PH$ & Element-wise & Unfused \\
        Scalar + low-rank & $P+LHR+PRH$ & Rank-$R$ & Unfused \\
        \bottomrule
    \end{tabular}
    \caption{Implementation differences among the evaluated projector families.}
    \label{tab:appendix_projector_implementations}
\end{table}

\subsection{Phase-Specific Routing and Folding}
\label{sec:appendix_routing_folding}

All retained sets and projector lookups below are defined separately for each
MoE layer; we omit the layer index for clarity. Let the original routes for token $x_i$ be
$S_K(x_i)=(e_{i,1},\ldots,e_{i,K})$, ordered by router score, and let
$w_{i,e}$ be the routed weight of expert $e$ and $h_e$ its cached
output-norm estimate; $w_{i,e}=0$ when $e\notin S_K(x_i)$. Let
$K_{\mathrm{pre}}$ be the per-token prefill budget, let $\mathcal{X}_q$ be the
tokens processed in decode step $q$, let $B=|\mathcal{X}_q|$, and let $D$ be
the decode expert-pool budget. The prefill path ranks the token's routed
experts by estimated contribution magnitude and retains
$K_{\mathrm{pre}}$ of them:
\begin{equation}
    B_{\mathrm{pre}}(x_i)
    =
    \operatorname{TopK}_{e\in S_K(x_i)}
    \left(w_{i,e}h_e,K_{\mathrm{pre}}\right).
    \label{eq:appendix_prefill_budget}
\end{equation}
Decode supports two retained-pool selectors that share the same folding
operator. The dynamic path used for the cross-model results and online
decode measurement ranks the union of routed experts by its aggregate
contribution:
\begin{equation}
    B_{\mathrm{dec}}^{\mathrm{dyn}}(\mathcal{X}_q)
    =
    \operatorname{TopK}_{e\in\cup_i S_K(x_i)}
    \left(\sum_{x_i\in\mathcal{X}_q}w_{i,e}h_e,D\right).
    \label{eq:appendix_decode_dynamic_budget}
\end{equation}
An additional static implementation removes this batch reduction from the
per-step decode path. It forms a pool once from the calibration-time norms,
\begin{equation}
    B_{\mathrm{dec}}^{\mathrm{static}}
    =
    \operatorname{TopK}_{e\in\{1,\ldots,E\}}(h_e,D),
    \label{eq:appendix_decode_static_budget}
\end{equation}
and precomputes one target and scalar for every possible source expert.
Given any retained set $B_\phi$, an omitted source expert selects the target
with minimum calibrated projection loss:
\begin{equation}
    \pi(s\mid B_\phi)
    =
    \arg\min_{t\in B_\phi}\ell_{s\to t},
    \label{eq:appendix_minimum_loss_assignment}
\end{equation}
where $B_\phi$ is the prefill set in Eq.~\ref{eq:appendix_prefill_budget}
or either decode pool in
Eqs.~\ref{eq:appendix_decode_dynamic_budget}--\ref{eq:appendix_decode_static_budget}.

\paragraph{Prefill folding.}
The prefill selector places the retained routes from Eq.~\ref{eq:appendix_prefill_budget} first, after which the Triton operator coalesces omitted routes assigned to the same retained target:
\begin{equation}
    \widetilde{w}_{i,t}
    =
    w_{i,t}
    +
    \sum_{\substack{s\in S_K(x_i)\setminus B_{\mathrm{pre}}(x_i)\\
                    \pi(s\mid B_{\mathrm{pre}}(x_i))=t}}
    w_{i,s}s_{s\to t}^{\star},
    \quad t\in B_{\mathrm{pre}}(x_i).
    \label{eq:appendix_prefill_fold}
\end{equation}
The operator does not materialize omitted expert outputs. Retained expert identifiers remain unchanged, so the fused MoE kernel executes only $K_{\mathrm{pre}}$ experts per token.

\paragraph{Decode remapping.}
Decode first obtains a dynamic or static pool. Each routed expert outside
that pool selects its target using
Eq.~\ref{eq:appendix_minimum_loss_assignment} and is remapped in place:
\begin{equation}
    (E_s,w_{i,s})
    \longmapsto
    \left(E_{\pi(s\mid B_{\mathrm{dec}})},
    w_{i,s}s_{s\to\pi(s\mid B_{\mathrm{dec}})}^{\star}\right).
    \label{eq:appendix_decode_remap}
\end{equation}
The lookup and remapping occur before dispatch, so every routed identifier presented to the fused kernel belongs to a pool of at most $D$ experts.

\paragraph{Decode kernels.}
Here $B_{\mathrm{dec}}$ is the dynamic or static pool selected above. The
evaluated dynamic path scans the $BK$ routed slots to build
Eq.~\ref{eq:appendix_decode_dynamic_budget}, then searches $D$ candidate
targets for each omitted route. The static implementation instead caches
    $\pi(s\mid B_{\mathrm{dec}})$ and
    $s_{s\to\pi(s\mid B_{\mathrm{dec}})}^\star$ for all $E$ sources.
Listing~\ref{lst:exfold_decode_kernel} is the resulting per-route kernel: one target
and scalar lookup per routed slot, or $O(BK)$ work, while preserving the
original $[B,K]$ layout.

\begin{listing}[t]
\begin{lstlisting}[
    language=Python,
    basicstyle=\scriptsize\ttfamily,
    numbers=none,
    xleftmargin=0pt,
    columns=fullflexible]
@triton.jit
def _static_projector_remap_kernel(
        weights, ids, target_ids, target_scale,
        out_weights, out_ids, B,
        K: tl.constexpr, BLOCK_B: tl.constexpr):
    batch = tl.program_id(0) * BLOCK_B \
            + tl.arange(0, BLOCK_B)
    valid = batch < B

    for k in tl.static_range(K):
        w = tl.load(weights + batch * K + k,
                    mask=valid)
        src = tl.load(ids + batch * K + k,
                      mask=valid)
        dst = tl.load(target_ids + src,
                      mask=valid, other=0)
        scale = tl.load(target_scale + src,
                        mask=valid, other=1.0)
        tl.store(out_weights + batch * K + k,
                 w * scale, mask=valid)
        tl.store(out_ids + batch * K + k,
                 dst, mask=valid)
\end{lstlisting}
\caption{Abbreviated static ExFold remapping kernel.}
\label{lst:exfold_decode_kernel}
\end{listing}

\section{Cross-Model Quality Results}
\label{sec:appendix_cross_model}

Table~\ref{tab:appendix_deepseek_results} reports DeepSeek-V2-Lite-Chat under joint compression. Table~\ref{tab:appendix_qwen35_results} separates prefill-only, decode-only, and joint folding on Qwen3.5-35B-A3B. In the method labels, P4 retains four routed experts per prefill token; D32 and D64 cap each decode batch's active expert pool at 32 and 64 experts; a combined label applies both constraints. For Qwen3.5 P4 and P4+D64, directed pairs with calibrated loss above 0.99 are excluded from target assignment; after prefill folding, the retained weights are rescaled per token to preserve the original routed-weight sum. This safeguard is not enabled for D64. All rows use complete benchmark coverage under the evaluation protocol described above.
The ``Eval+'' column reports HumanEval+ pass@1.

\begin{table}[!ht]
    \centering
    \small
    \setlength{\tabcolsep}{3.2pt}
    \resizebox{\columnwidth}{!}{%
    \begin{tabular}{lccccc}
        \toprule
        \rowcolor{gray!30}
        \textbf{Method} & \textbf{IFEval} & \textbf{GPQA} &
        \textbf{Eval+} & \textbf{MMLU$_{pro}$} &
        \textbf{Avg.} \\
        \midrule
        Original Top-6 & \textbf{42.14} & 23.23 & 46.34 & \textbf{26.52} & 34.56 \\
        Direct Top-4 & 40.85 & 24.24 & 46.95 & 24.71 & 34.19 \\
        \rowcolor{blue!5}
        \textbf{\sys P4+D32} & 41.96 & \textbf{26.77} & \textbf{48.17} & 24.75 & \textbf{35.41} \\
        \bottomrule
    \end{tabular}}
    \caption{DeepSeek-V2-Lite-Chat quality; bold marks the column best.}
    \label{tab:appendix_deepseek_results}
\end{table}

\begin{table}[!ht]
    \centering
    \small
    \setlength{\tabcolsep}{2.8pt}
    \resizebox{\columnwidth}{!}{%
    \begin{tabular}{lcccccc}
        \toprule
        \rowcolor{gray!30}
        \textbf{Method} & \textbf{MATH$_{500}$} & \textbf{IFEval} &
        \textbf{IFBench} & \textbf{GPQA} & \textbf{Eval+} &
        \textbf{Avg.} \\
        \midrule
        Original Top-8 & 89.00 & \textbf{87.06} & 29.20 & 82.83 & 88.41 & 75.30 \\
        Direct Top-4 & 32.80 & 81.52 & 33.53 & 82.32 & 87.20 & 63.47 \\
        \cmidrule(lr){1-7}
        \rowcolor{blue!5}
        \textbf{\sys P4} & 88.40 & 82.62 & 32.59 & 82.32 & \textbf{92.68} & 75.72 \\
        \rowcolor{blue!5}
        \textbf{\sys D64} & \textbf{89.20} & 85.21 & \textbf{38.40} & 81.31 & 89.63 & 76.75 \\
        \rowcolor{blue!5}
        \textbf{\sys P4+D64} & 88.40 & 83.36 & 33.77 & \textbf{86.36} & 92.07 & \textbf{76.79} \\
        \bottomrule
    \end{tabular}}
    \caption{Qwen3.5-35B-A3B quality; bold marks the column best.}
    \label{tab:appendix_qwen35_results}
\end{table}

\section{Expert Geometry and Projector Structure}
\label{sec:appendix_analysis}

\subsection{Full-Layer Expert Geometry}
\label{sec:appendix_geometry}

We visualize expert-output geometry for every MoE layer. For Qwen3 output
matrices $\mathbf{E}_s$ and $\mathbf{E}_t$ collected on common inputs, raw
similarity is
$1-\|\mathbf{E}_s-\mathbf{E}_t\|_F/d_{\max}$, where $d_{\max}$ is the largest
raw pairwise distance in that layer. For this diagnostic, the aligned view fits
the closed-form scalar on the same captured outputs, averages the two directed
similarities, and reuses the same $d_{\max}$. GLM reconstructs the corresponding
raw and aligned distances from source-norm-weighted co-routing statistics and
averages the two directions using their observation counts. Thus, each aligned
distance is no larger than its raw counterpart; these analysis-only fits are
separate from the fixed deployment projectors visualized below. The magnitude bar charts show every
layer--expert position, use an independent y-axis in each layer, and highlight
the minimum- and maximum-magnitude experts. Unavailable observations are gray;
the isolated GLM layer-45 outlier is hatched and excluded from that panel's
axis scaling. Across observed
directed off-diagonal pairs, scalar
alignment raises the mean similarity from 0.449 to 0.613 for Qwen3. Across
observed symmetric GLM entries, the corresponding mean rises from 0.899 to
0.932 for GLM-4.5-Air.

\FloatBarrier

\begin{figure*}[p]
    \centering
    \includegraphics[width=0.95\textwidth]{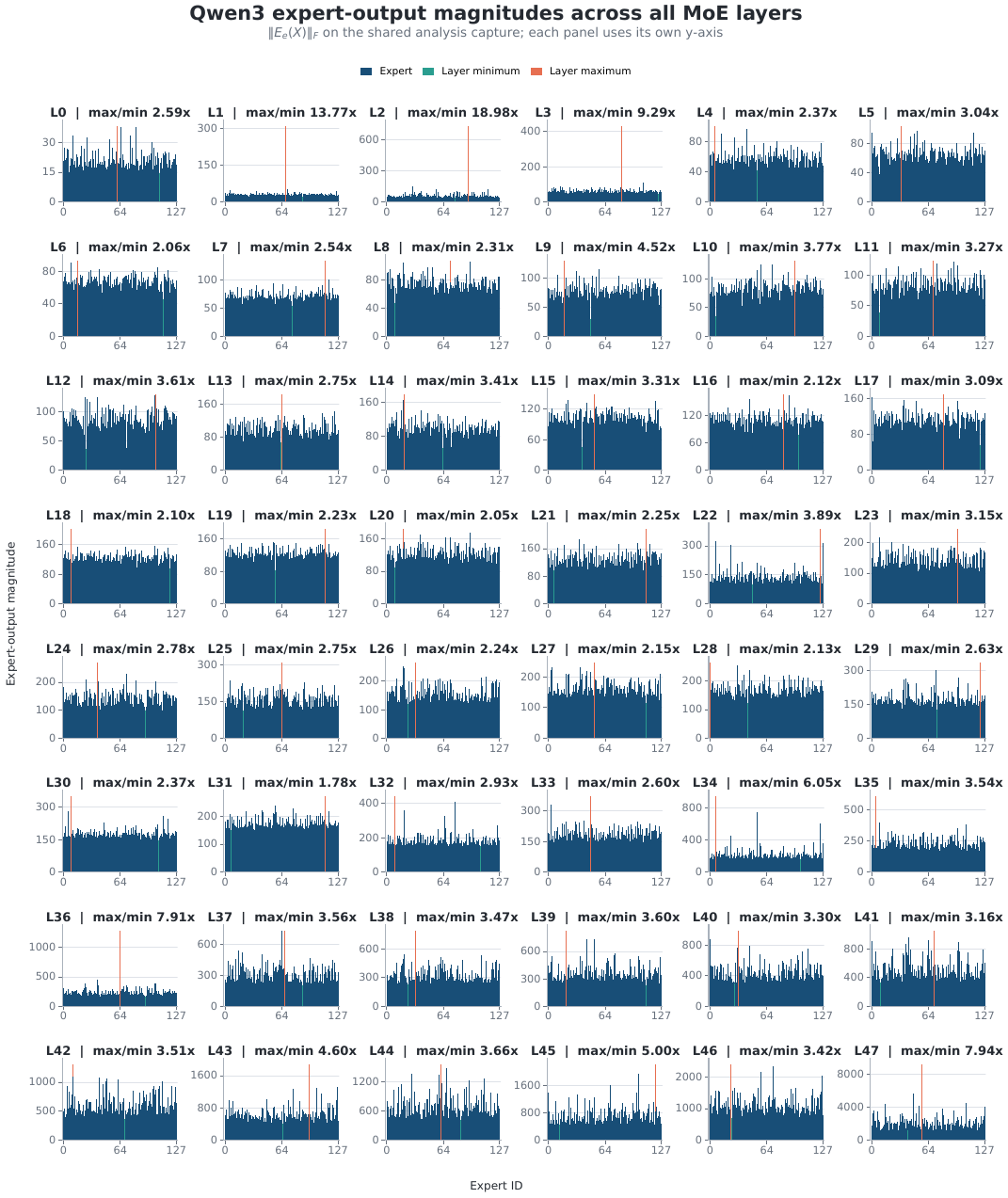}
    \caption{Expert-output magnitudes across all Qwen3 MoE layers.}
    \label{fig:appendix_norms}
\end{figure*}

\begin{figure*}[p]
    \ContinuedFloat
    \centering
    \includegraphics[width=0.95\textwidth]{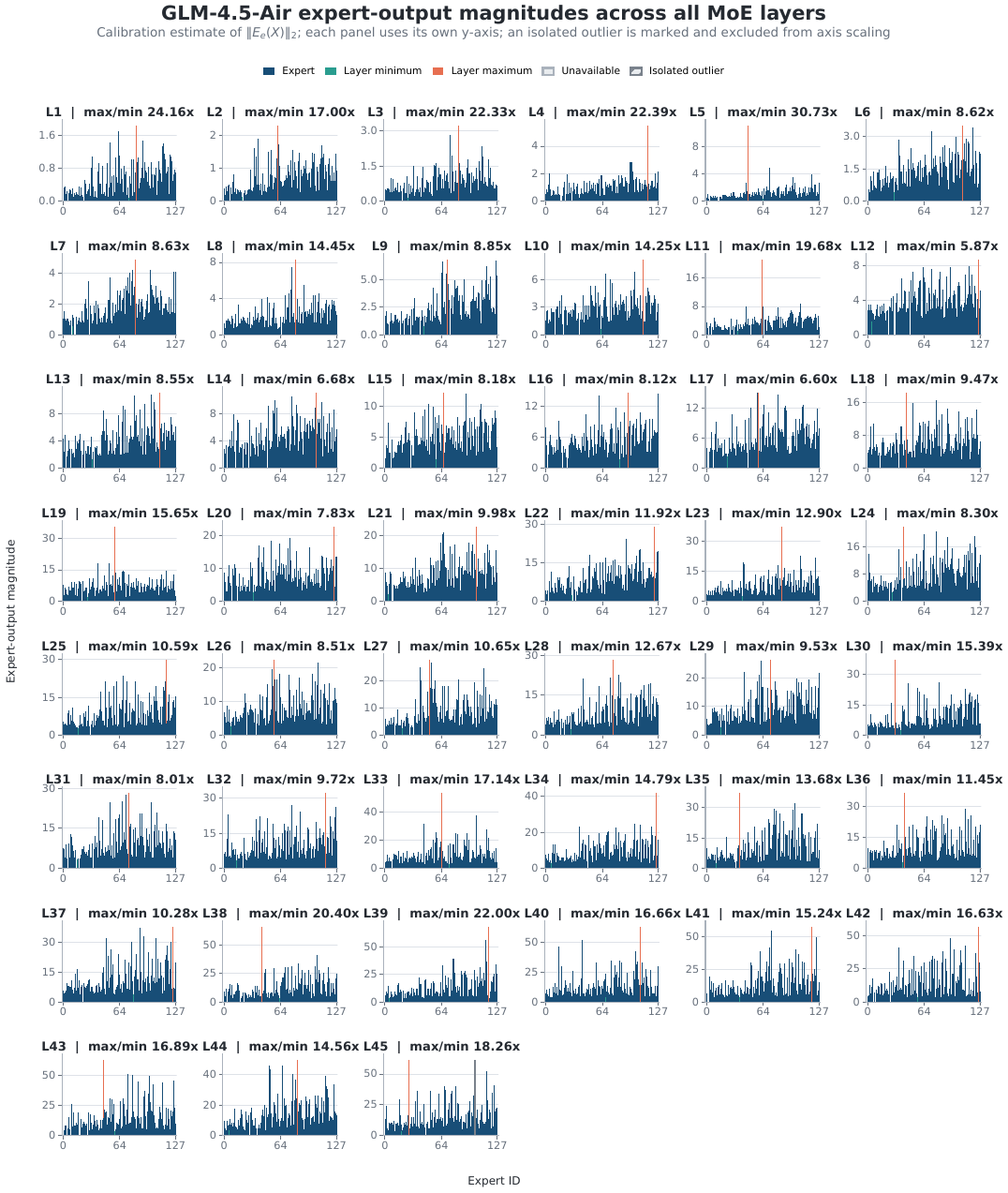}
    \caption{GLM-4.5-Air expert-output magnitudes (continued); hatching marks one isolated outlier.}
    \label{fig:appendix_glm_norms}
\end{figure*}

\begin{figure*}[p]
    \centering
    \includegraphics[width=0.98\textwidth]{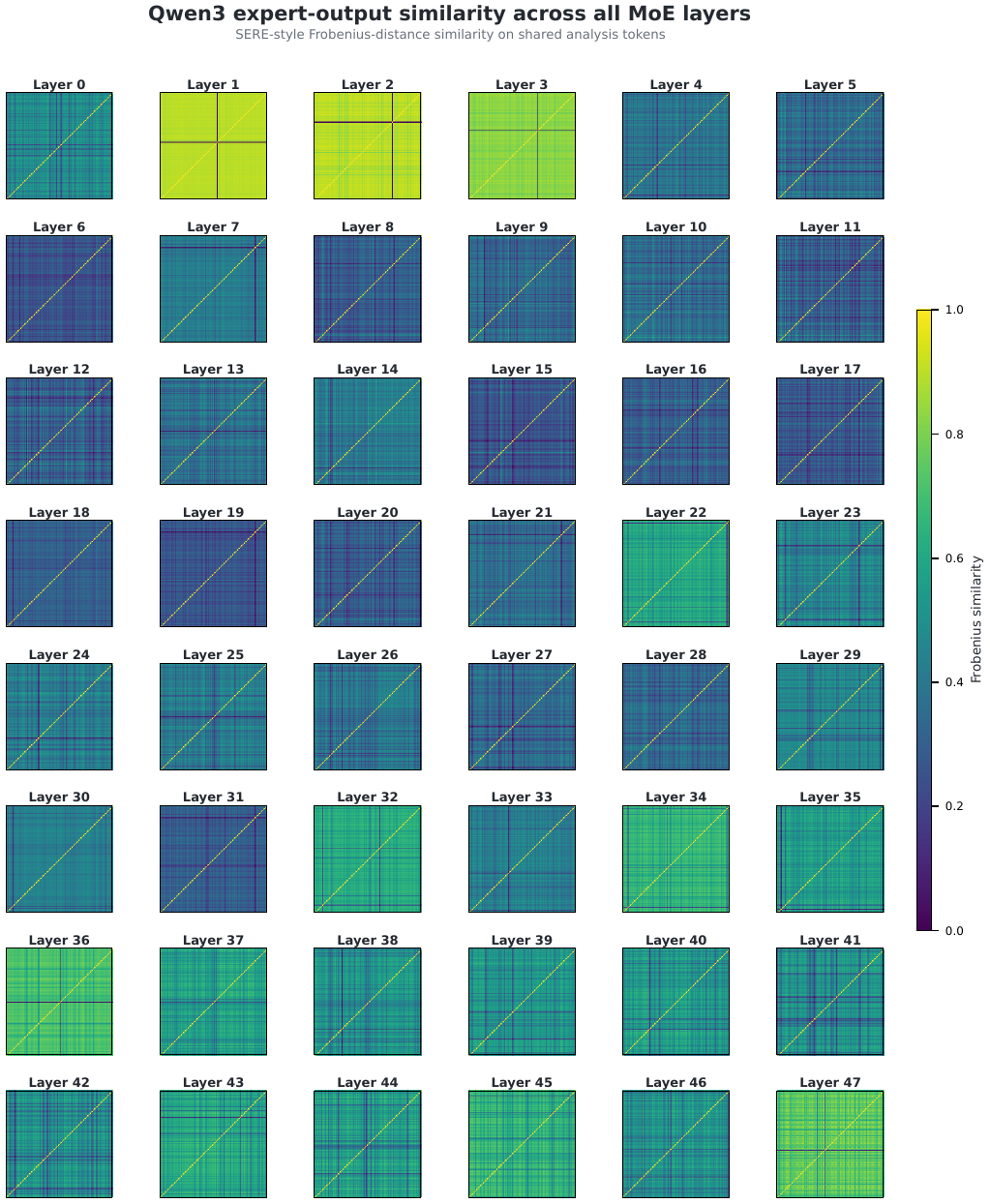}
    \caption{Raw expert-output similarity across all Qwen3 MoE layers.}
    \label{fig:appendix_qwen_raw_similarity}
\end{figure*}

\begin{figure*}[p]
    \centering
    \includegraphics[width=0.98\textwidth]{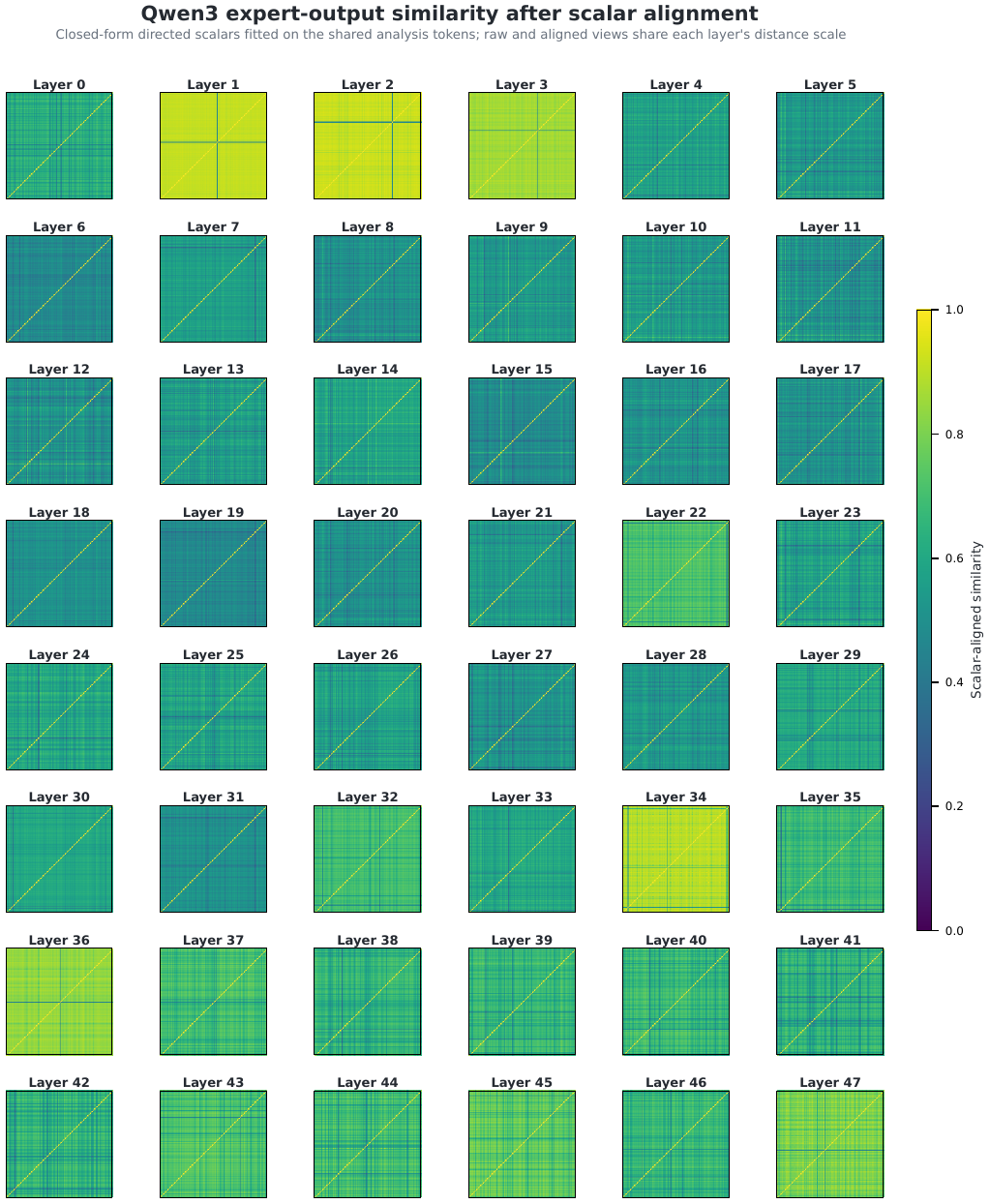}
    \caption{Scalar-aligned similarity for observed Qwen3 expert pairs.}
    \label{fig:appendix_qwen_scalar_similarity}
\end{figure*}

\paragraph{Qwen3-30B-A3B.}
Figures~\ref{fig:appendix_norms}, \ref{fig:appendix_qwen_raw_similarity}, and~\ref{fig:appendix_qwen_scalar_similarity} show all 48 layers and 128 routed experts. The matrices exhibit layer-dependent structure, and the norm view confirms expert-wise magnitude variation.

\begin{figure*}[p]
    \centering
    \includegraphics[width=0.98\textwidth]{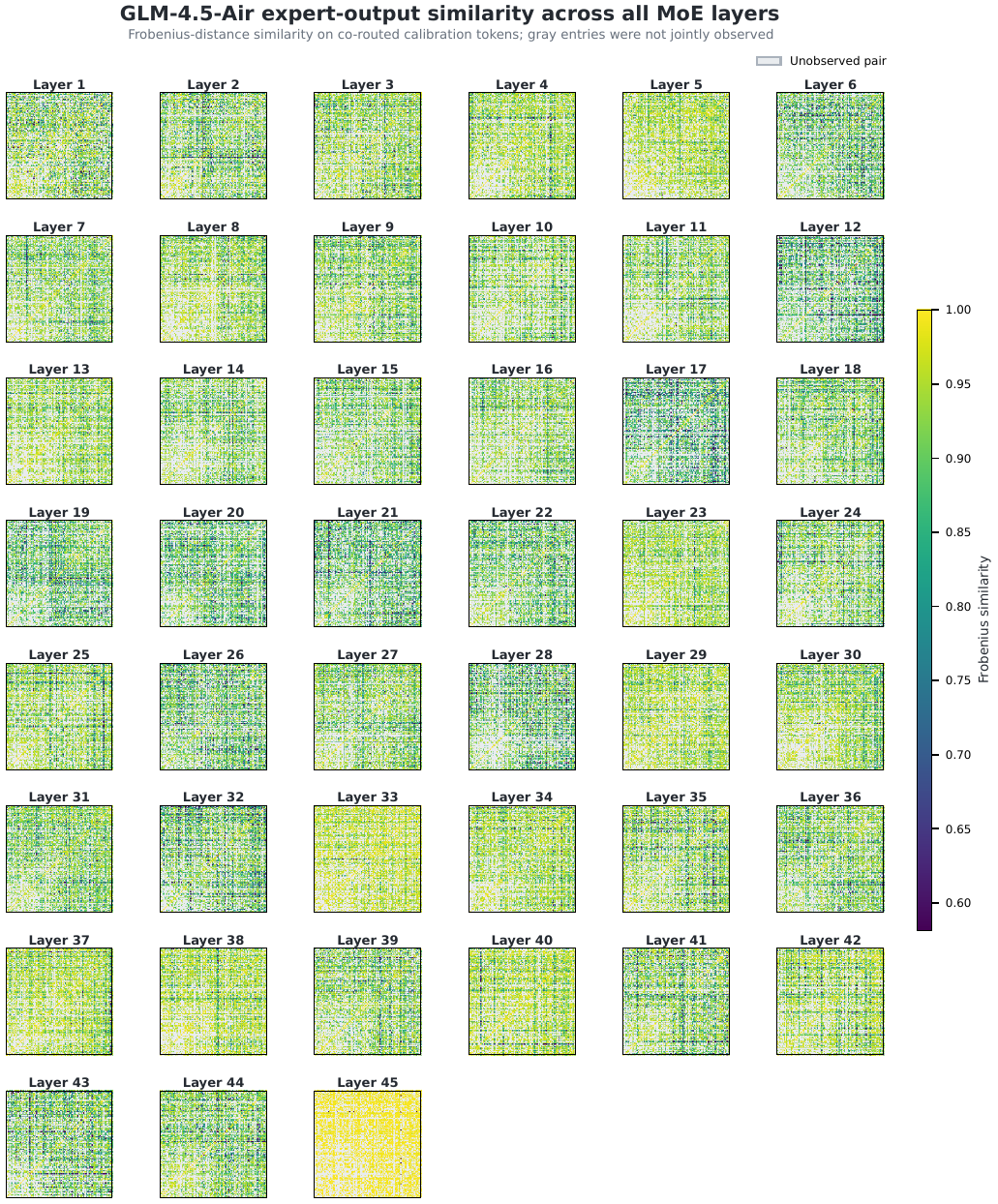}
    \caption{Raw co-routed expert similarity across all GLM-4.5-Air MoE layers.}
    \label{fig:appendix_glm_raw_similarity}
\end{figure*}

\begin{figure*}[p]
    \centering
    \includegraphics[width=0.98\textwidth]{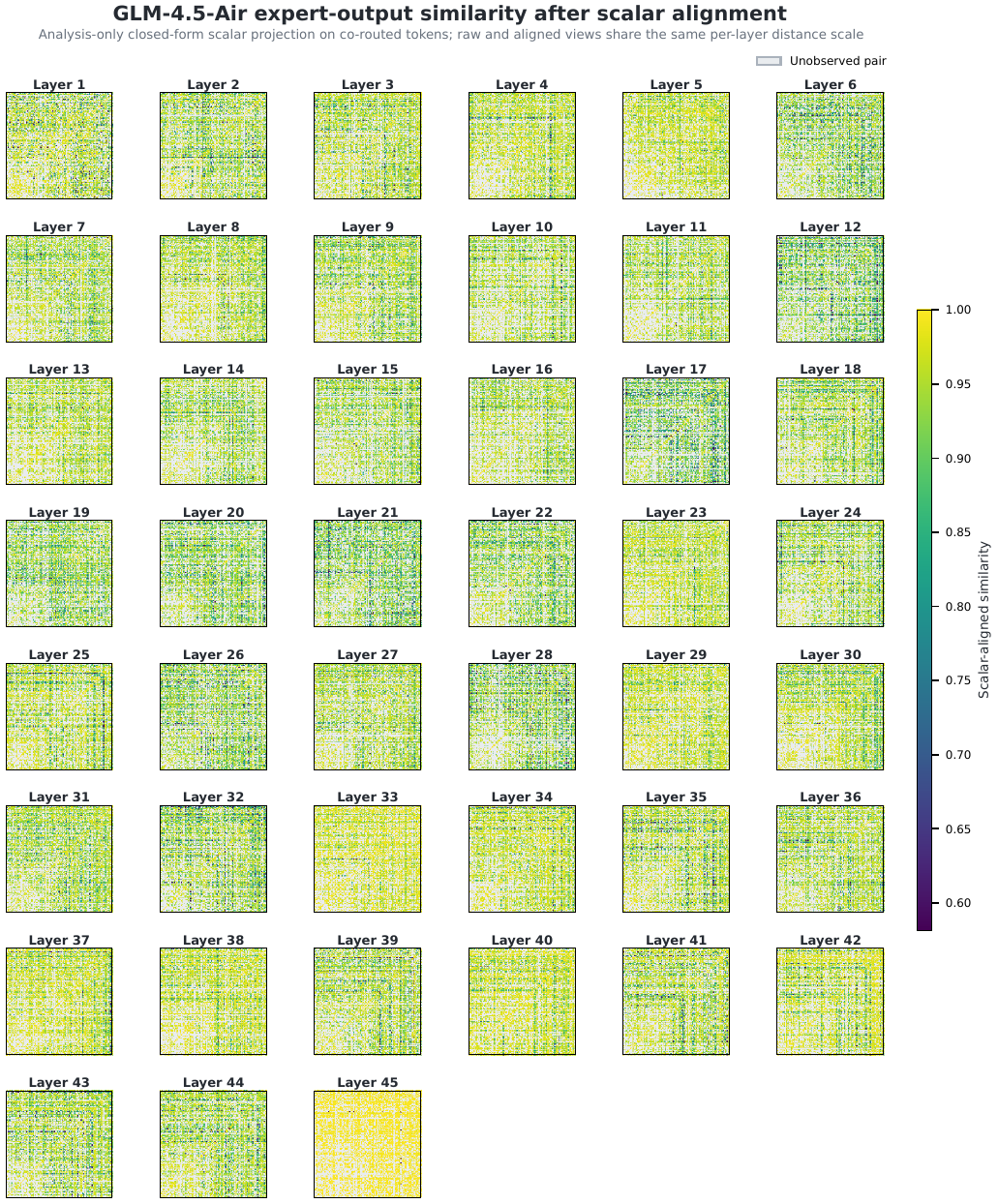}
    \caption{Scalar-aligned co-routed expert similarity across all GLM-4.5-Air MoE layers.}
    \label{fig:appendix_glm_scalar_similarity}
\end{figure*}

\paragraph{GLM-4.5-Air.}
Figures~\ref{fig:appendix_glm_norms}, \ref{fig:appendix_glm_raw_similarity}, and~\ref{fig:appendix_glm_scalar_similarity} provide the corresponding 45-layer diagnostics. GLM statistics are collected for co-routed expert pairs; gray cells denote pairs that were never jointly observed. The matrices again show nonuniform pair structure and output-magnitude differences.

\FloatBarrier

\begin{figure*}[p]
    \centering
    \begin{minipage}{0.96\textwidth}
    \subsection{Layer-Wise Projector Structure}
    \label{sec:appendix_projector_structure}
    Figures~\ref{fig:appendix_qwen_scalars} and~\ref{fig:appendix_qwen_losses} visualize dense common-input projector geometry for Qwen3. Every expert processes the same hidden-state matrix $X$; rows are source experts, columns are candidate targets, and self-pairs are suppressed. Scalar colors are clipped symmetrically at the 99th percentile of $|s_{s\to t}^{\star}|$. For a layer $l$, Figure~\ref{fig:appendix_qwen_losses} reports the directed normalized Frobenius loss
    $\widetilde{\ell}_{s\to t}=\|E_s(X)-s_{s\to t}^{\star}E_t(X)\|_F/d_{\max}^{(l)}=1-S_{\mathrm{scalar}}(s,t)$,
    where $d_{\max}^{(l)}$ is the maximum raw pair distance in that layer. Its median over directed off-diagonal pairs is 0.407, and 79.9\% of pairs are below 0.5. This common-input diagnostic avoids conflating missing pair support with projection quality; lower values indicate more compatible transfers.
    \end{minipage}
    \vspace{1mm}
    \includegraphics[width=0.88\textwidth]{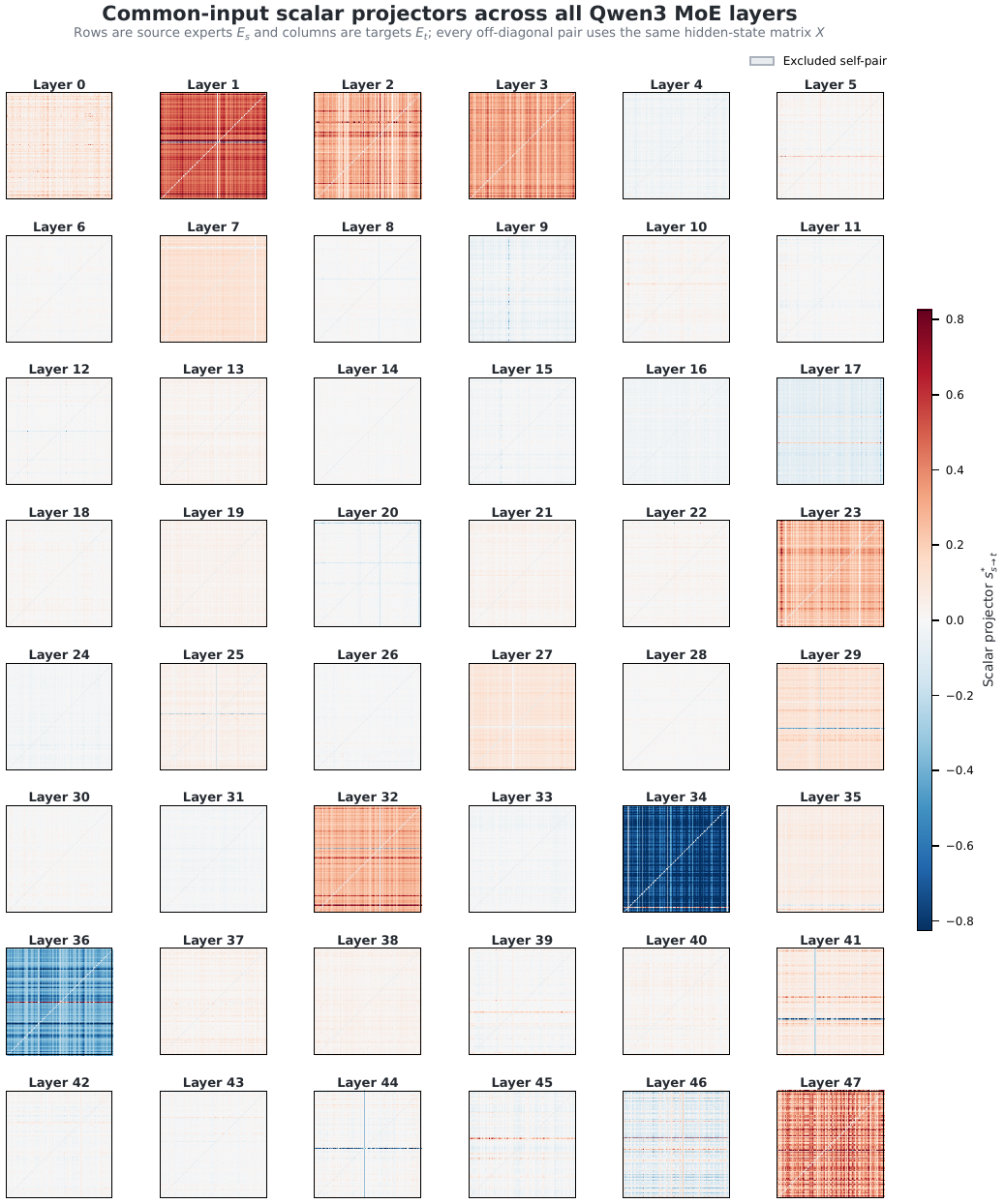}
    \caption{Common-input scalar-projector matrices across all Qwen3 MoE layers.}
    \label{fig:appendix_qwen_scalars}
\end{figure*}

\begin{figure*}[p]
    \centering
    \includegraphics[width=0.88\textwidth]{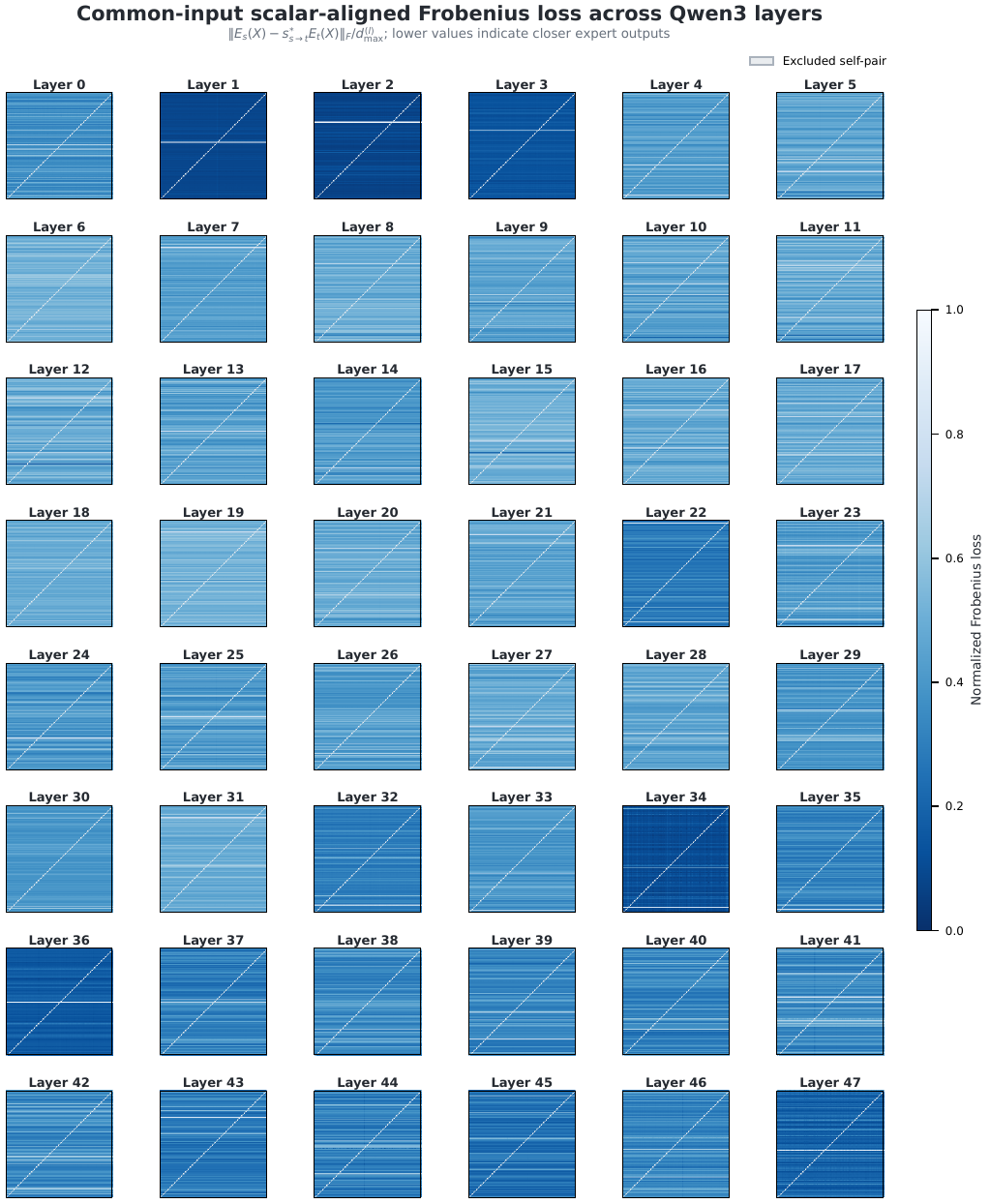}
    \caption{Common-input scalar-aligned Frobenius losses across all Qwen3 MoE layers.}
    \label{fig:appendix_qwen_losses}
\end{figure*}

\FloatBarrier

\begin{figure*}[p]
    \centering
    \begin{minipage}{0.96\textwidth}
    \section{Token-Level Reconstructions}
    \label{sec:appendix_token_cases}
    Figure~\ref{fig:appendix_token_cases} provides proxy-space illustrations
    for tokens from mathematics, scientific reasoning, instruction following, and
    code generation. The columns compare the original Top-8 routes with Direct
    Top-4, MoDES token-wise selection~\cite{huang2025modes}, REAP using Top-8
    within a static 64-expert pool~\cite{lasby2025reap}, and ExFold Top-4.
    In this comparison, MoDES applies its calibrated layer-wise threshold to
    retain a token-dependent subset of at most four leading routes.
    Expert directions come from two-dimensional classical multidimensional
    scaling (MDS) of the
    all-expert cosine-distance matrix, and arrow lengths equal router weight times the
    calibrated expert-output norm; segment labels $1,\ldots,8$ denote
    route-rank positions rather than global expert identifiers, and primed
    labels denote REAP routes. ExFold omits per-segment labels for clarity.
    Colored arrows show individual expert contributions in a strict
    head-to-tail chain: every arrow starts at the endpoint of the preceding
    contribution, and the chain endpoint is the corresponding aggregate.
    Background arrows ending in diamonds show the original aggregate
    $\mathbf{y}$ and each method's approximate aggregate
    $\widehat{\mathbf{y}}$; the latter is dashed. The displayed proxy distance
    is their relative Euclidean distance in this two-dimensional space.
    Within each dataset, we inspect at most eight evaluation examples and 96
    deterministically spaced tokens per example. We select an illustrative case
    whose Direct-Top-4 proxy distance is at least 0.15, for which ExFold
    reduces that distance by at least 15\%, and for which ExFold has the
    smallest proxy distance among the displayed approximations. To span model
    depth, the four datasets target layers 0, 16, 32, and 47, respectively;
    each panel uses the nearest qualifying layer and the first case in
    deterministic scan order. These diagrams illustrate routing geometry;
    their reported distances are proxy-space quantities and are not hidden-state
    reconstruction errors. They are qualitative examples rather than aggregate
    evaluation evidence.
    \end{minipage}
    \includegraphics[width=0.78\textwidth]{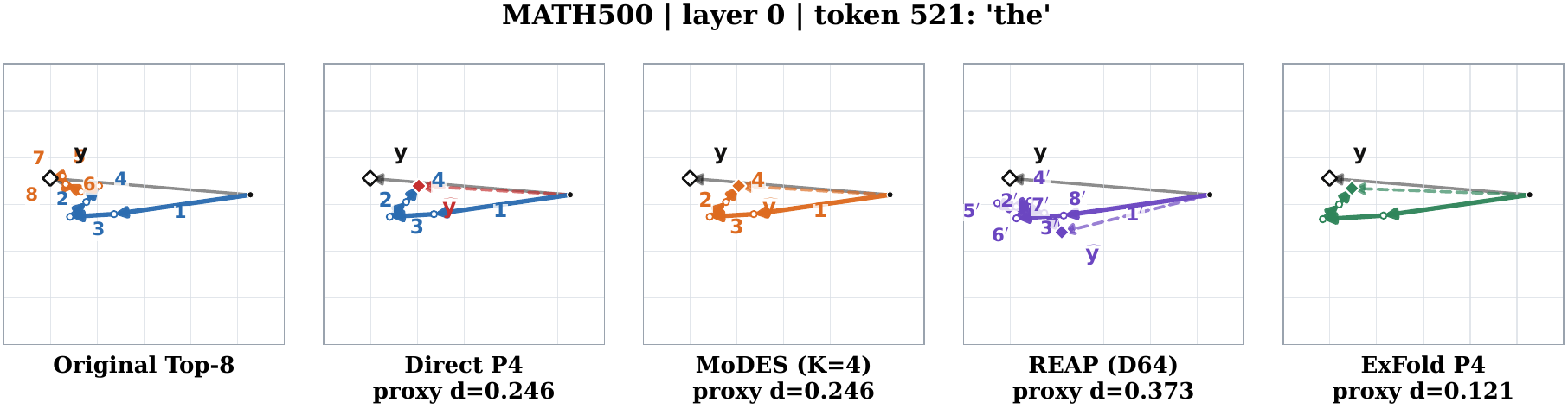}
    \vspace{0.3mm}
    \includegraphics[width=0.78\textwidth]{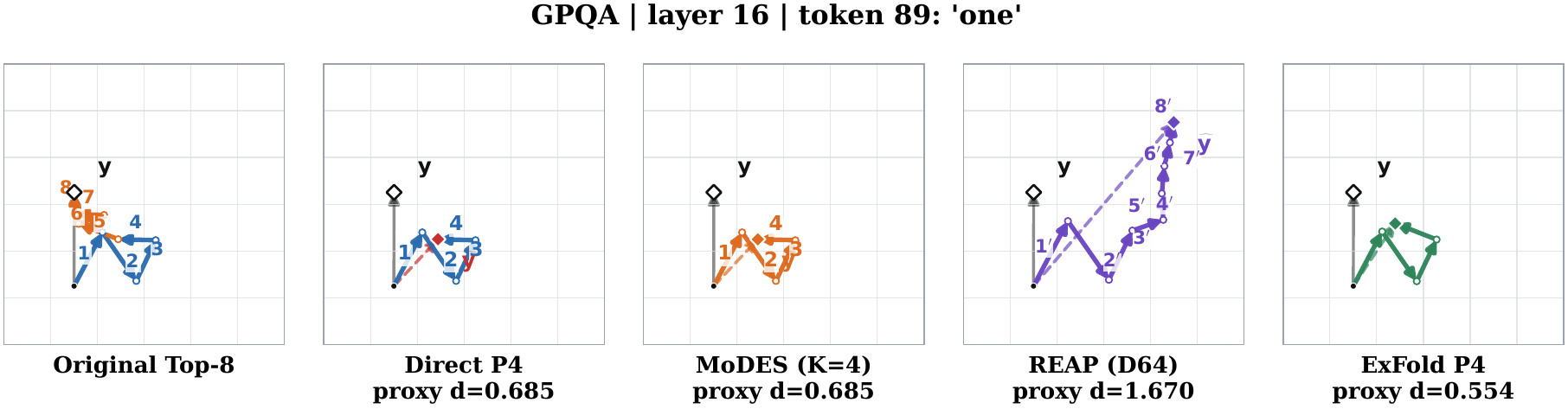}
    \vspace{0.3mm}
    \includegraphics[width=0.78\textwidth]{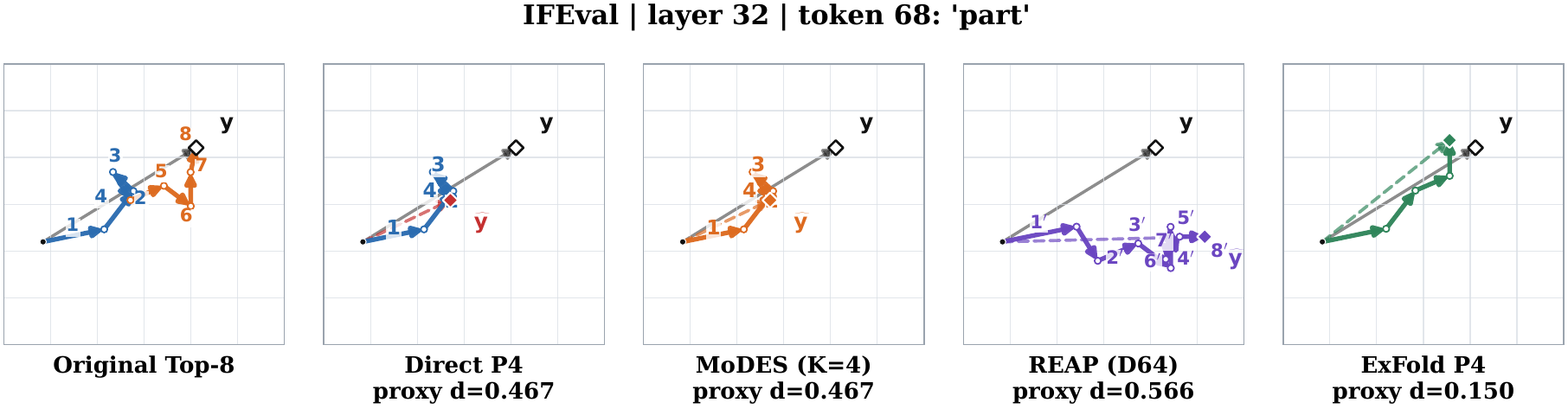}
    \vspace{0.3mm}
    \includegraphics[width=0.78\textwidth]{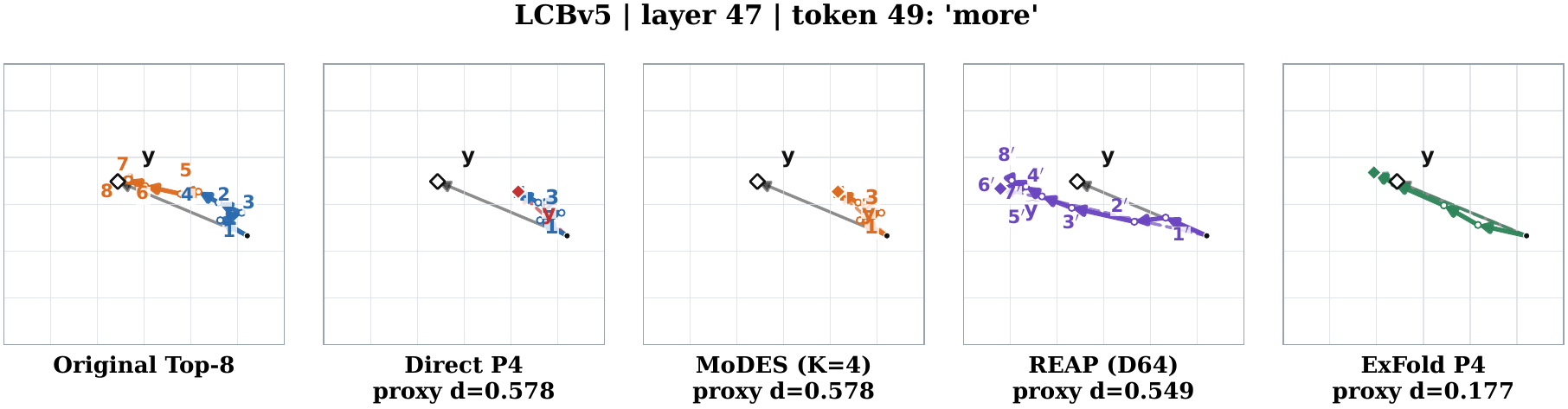}
    \caption{Token-level proxy reconstructions. Expert contributions are accumulated strictly head-to-tail; each chain terminates at $\mathbf{y}$ or $\widehat{\mathbf{y}}$.}
    \label{fig:appendix_token_cases}
\end{figure*}

    \section{DeepSeek-V4-Flash Extension Details}
\label{sec:dsv4_repro}

\paragraph{Model and quality runtime.}
DeepSeek-V4-Flash contains 284B total and 13B activated parameters, 43
transformer layers, 256 routed experts plus one shared expert, and Top-6
routing~\cite{deepseekv4}. Quality evaluation uses tensor parallelism eight,
\texttt{max\_num\_seqs=32}, and CUDA Graph. Every method is evaluated with
the same prompts, chat formatting, decoding rules, and scorers. The full-suite
collector is required to have no missing tasks or infrastructure failures
before it is admitted to Table~\ref{tab:dsv4_quality}. All averages and
retention ratios are computed from underlying unrounded task scores.


\paragraph{Calibration data and disclosure.}
The released DeepSeek matrix is calibrated from 64 unlabeled inputs: 56
general instruction, code, and mathematics inputs from Tulu-3, plus eight
benchmark inputs (four IFEval and four IFBench) without labels, reference
answers, or evaluator feedback. At most 64 observer tokens are collected per
sequence, with maximum input length 4,096. This is
\emph{transductive, benchmark-aware calibration}, not a zero-contact held-out
evaluation. Calibration uses no benchmark answers and changes no model
weights, but the input exposure must be preserved when reporting the results.

The DeepSeek artifact uses common-input expert outputs, source-output-norm
weighting, and unbounded least-squares scalars. Unlike the Qwen configuration,
it does not clip scalar coefficients.

\paragraph{Confidence-aware P3 safeguard.}
DeepSeek-V4-Flash exhibits a wider range of router scores and expert-output
magnitudes than the primary Qwen model. For an omitted source $s$ and retained
target $t$, we therefore compute a source-relative residual
\begin{equation}
    \begin{aligned}
        \ell_{s\rightarrow t}^{\mathrm{rel}}
        &=
        \frac{\sum_i w_i
        \|\mathbf{u}_i-a_{s\rightarrow t}\mathbf{v}_i\|_2^2}
        {\sum_i w_i\|\mathbf{u}_i\|_2^2+\epsilon}, \\
        c_s&=\left[1-\min_t
        \ell_{s\rightarrow t}^{\mathrm{rel}}\right]_0^1 .
    \end{aligned}
    \label{eq:dsv4_confidence}
\end{equation}
The scalar-transfer part $w_s c_s a_{s\rightarrow t}$ is added to the
minimum-loss target. The remaining weight $w_s(1-c_s)$ falls back to the
retained Top-3 routes in proportion to their original router weights. This
continuous safeguard does not execute additional experts: P3 still invokes
exactly three routed experts. It only avoids forcing a poorly calibrated pair
to absorb the entire omitted contribution.

\paragraph{Speed protocol and operating-point boundary.}
All DeepSeek speed measurements use one H800 server with model weights, code,
data, and outputs on local storage. The vLLM runtime uses tensor parallelism
four and CUDA Graph. Online load curves use \texttt{max\_num\_seqs=32} and
the same request schedule for every method. Prefill uses 8,192 input tokens
and one output token at QPS 1, 2, 4, 6, and 8. Decode uses one input token and
256 output tokens at QPS 2, 4, 8, 12, 16, 24, and 32. The offline-throughput
panel uses \texttt{max\_num\_seqs=128}, QPS 64, concurrency 128, 32 warmups,
and 1,024 measured requests.

The TTFT peak at QPS 4 includes queueing amplification and is therefore a
serving-level speedup rather than a pure-kernel claim. Under the matched
\texttt{max\_num\_seqs=32} decode limit, D64 stabilizes near 1.15$\times$
TPOT at QPS 8--32, while D128 remains near parity. The saturated
\texttt{max\_num\_seqs=128} run yields a 1.286$\times$ output-throughput gain
for D64. Because quality is measured with \texttt{max\_num\_seqs=32}, the
throughput bar is reported as a separate saturation boundary rather than the
same quality--speed operating point.

\paragraph{Open-source reproduction.}
The repository at \url{https://github.com/Time-Rune/ExFold-MoE} contains the Qwen3 and
DeepSeek-V4-Flash runtime patches, CUDA/Triton kernels, final calibration
matrices, correctness tests, and one-command quality and speed launchers.
Model weights and benchmark datasets remain under their original licenses and
are downloaded from their providers.


\end{document}